\pdfoutput=1

\documentclass[11pt]{article}

\usepackage[final]{acl}
\usepackage[inline]{enumitem}
\usepackage{colortbl}
\usepackage{amsmath}
\usepackage{booktabs}
\usepackage{multirow}
\usepackage{times}
\usepackage{latexsym}
\usepackage[T1]{fontenc}

\usepackage[utf8]{inputenc}

\usepackage{microtype}

\usepackage{inconsolata}

\usepackage{graphicx}
\usepackage{amsfonts}
\usepackage{algorithm}
\usepackage{algpseudocode}
\usepackage{tabularx}
\usepackage{threeparttable}
\usepackage{pifont}

\definecolor{pale_green}{HTML}{ECF5E4}
\definecolor{mygreen}{HTML}{ECF5E4}

\title{Safety-Utility Conflicts Are Not Global:\\ Surgical Alignment via Head-Level Diagnosis}
\author{
    Wang Cai$^{1,2}$, 
    Yilin Wen$^{1}$,
    Jinchang Hou$^{1}$, 
    Du Su$^{4}$,
    Guoqiu Wang$^{1}$,\\
    Zhonghou Lv$^{1, \dagger}$, 
    Chenfu Bao$^{1, \dagger}$, 
    Yunfang Wu$^{3, \dagger}$\thanks{
        $^{\dagger}$ Corresponding authors. 
        Other emails: \texttt{\{wenyilin, houjinchang, wangguoqiu\}@baidu.com}, \texttt{sudu@ict.ac.cn}
    } \\
    $^{1}$Baidu Inc. \\
    $^{2}$School of Software and Microelectronics, Peking University \\ 
    $^{3}$School of Computer Science, Peking University \\
    $^{4}$State Key Laboratory of AI Safety, Institute of Computing Technology, CAS \\
    \texttt{caiwang@stu.pku.edu.cn}, 
    \texttt{\{lvzhonghou, baochenfu\}@baidu.com},
    \texttt{wuyf@pku.edu.cn}
}

\begin{document}
\maketitle
\begin{abstract}
Safety alignment in Large Language Models (LLMs) inherently presents a multi-objective optimization conflict, often accompanied by an unintended degradation of general capabilities. Existing mitigation strategies typically rely on global gradient geometry to resolve these conflicts, yet they overlook \textit{Modular Heterogeneity} within Transformers, specifically that the functional sensitivity and degree of conflict vary substantially across different attention heads. Such global approaches impose uniform update rules across all parameters, often resulting in suboptimal trade-offs by indiscriminately updating utility sensitive heads that exhibit intense gradient conflicts. To address this limitation, we propose \textbf{Conflict-Aware Sparse Tuning (CAST)}, a framework that integrates head-level diagnosis with sparse fine-tuning. CAST first constructs a pre-alignment conflict map by synthesizing \textit{Optimization Conflict} and \textit{Functional Sensitivity}, which then guides the selective update of parameters. Experiments reveal that alignment conflicts in LLMs are not uniformly distributed. We find that the drop in general capabilities mainly comes from updating a small group of ``high-conflict'' heads. By simply skipping these heads during training, we significantly reduce this loss without compromising safety, offering an interpretable and parameter-efficient approach to improving the safety-utility trade-off.
\end{abstract}

\begin{figure}[t]
    \centering
    \captionsetup{font=footnotesize}
    \includegraphics[width=0.95\linewidth]{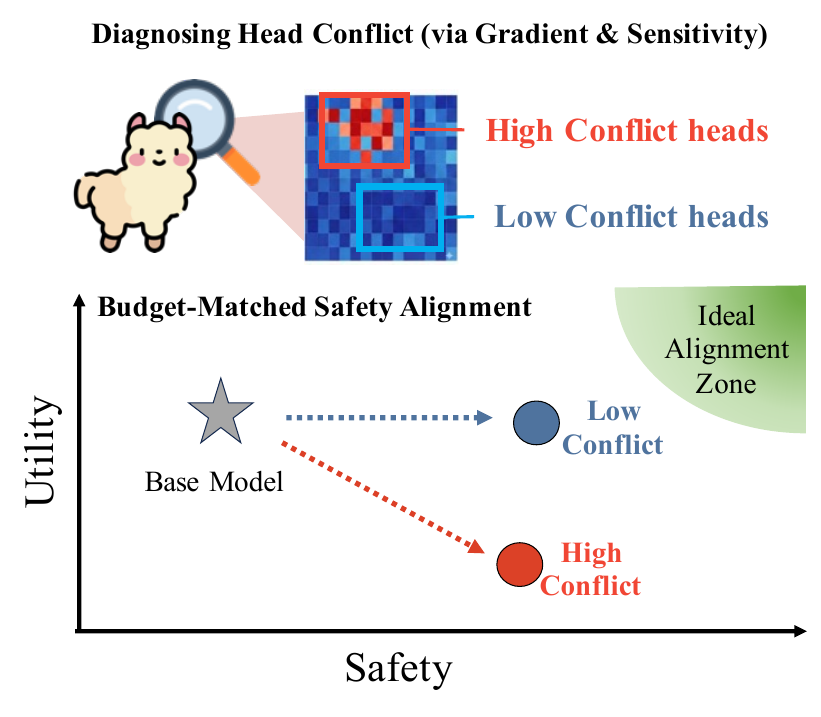}
    \caption{\small
    \textbf{Structural Localization of Alignment Conflict.} (Top) The diagnostic map reveals that conflict is concentrated in specific heads (red) rather than globally diffuse. (Bottom) Intervening on these High-Conflict heads precipitates utility degradation (red path), whereas CAST (blue path) selectively targets Low-Conflict heads to secure safety while preserving general capabilities.}
\label{fig:teaser}
\end{figure}

\section{Introduction}

Safety alignment is a prerequisite for deploying Large Language Models (LLMs), yet it frequently incurs an unintended degradation of general capabilities, widely known as the ``alignment tax''~\cite{ouyang2022training, touvron2023llama2}. Prevailing mitigation strategies typically frame this trade-off as a global Multi-Objective Optimization (MOO) problem. Researchers have employed geometric techniques, such as projecting conflicting gradients (e.g., PCGrad)~\cite{yu2020gradient} or balancing update directions (e.g., CAGrad)~\cite{liu2024conflict}, to resolve interference. However, these methods implicitly treat the Transformer as a homogeneous entity, applying uniform update rules across all parameters.

This global perspective overlooks a critical characteristic of Transformers: \textit{Modular Heterogeneity}~\cite{voita2019analyzing, michel2019sixteen}. Research confirms that attention heads are functionally specialized~\cite{clark2019doesbert, vig2019multiscale}, where specific attention heads mediate safety refusal~\cite{zhou2025roleattentionheadslarge}, whereas distinct subsets serve as critical hubs for reasoning~\cite{wang2022interpretability}. In such a system, relying solely on gradient geometry is insufficient. A functionally sensitive head, characterized by the fact that even minor modifications to its parameters can drastically alter model behavior, may suffer severe utility loss even under small updates. We argue that mitigating alignment cost requires moving beyond geometric techniques alone by explicitly coupling them with functional sensitivity to identify the \textbf{high-conflict heads} that are most susceptible to utility collapse.

In this study, we shift the analytical lens from global parameters to individual attention heads. We hypothesize that utility degradation is mainly caused by updating a small subset of heads that exhibit both high optimization conflict and functional sensitivity. To address this, we propose \textbf{Conflict-Aware Sparse Tuning (CAST)}. This framework employs \textbf{Head-Level Conflict Diagnosis} to construct a pre-alignment ``Conflict Map'' by synthesizing these two metrics, allowing us to pinpoint the specific units responsible for utility loss.

Guided by this diagnostic foundation, CAST executes a parameter-efficient alignment strategy. As illustrated in Figure~\ref{fig:teaser}, our framework reveals that alignment conflicts are \textbf{structurally localized} rather than globally diffuse. Through controlled safety alignment experiments with matched parameter budgets, CAST demonstrates that the choice of intervention region critically affects the outcome: updating "High-Conflict" heads incurs a severe utility penalty, whereas targeting ``Low-Conflict'' heads secures equivalent safety gains while significantly minimizing utility degradation.

Our contributions are summarized as follows:
\begin{itemize}[leftmargin=*,itemsep=2pt, topsep=0pt, parsep=0pt]
    \item \textbf{Structural Localization of Conflict:} We show that alignment conflicts are not spread across the whole model but are concentrated in specific attention heads. This offers a new modular perspective on the safety-utility trade-off.
    
    \item \textbf{The CAST Framework:} We propose CAST, which turns interpretability into a practical tool. It uses a Conflict Diagnosis to detects specific heads suitable for safety alignment, effectively avoiding the loss of general capabilities.
    
    \item \textbf{Predictive Validity \& Pareto Efficiency:} We verify that our diagnostic score accurately predicts performance drops (Pearson $r \in [0.73, 0.95]$). Guided by this, CAST achieves a better trade-off (Pareto frontier), preserving capabilities without sacrificing safety.
\end{itemize}

\section{Related Work}

\paragraph{Mitigating the Alignment-Utility Trade-off.}
Standard alignment algorithms frequently exact a ``tax'' on reasoning capabilities~\cite{lin2024mitigatingalignmenttaxrlhf}. 
Current mitigation strategies primarily operate at two levels: 
(1) \textbf{Training Process Level}, which seeks to reconcile objective conflicts via data engineering (e.g., replay or evolutionary prompts~\cite{dong2024abilities,xu2025wizardlm}) or gradient geometry manipulation (e.g., PCGrad~\cite{yu2020gradient}); 
(2) \textbf{Parameter Level}, which mitigates interference by constraining updates to low-rank subspaces (e.g., LoRA~\cite{hu2021lora}) or merging task vectors post-hoc (e.g., TIES-Merging~\cite{yadav2023ties}). 
However, both paradigms typically treat model parameters as homogeneous blocks or apply heuristic sparsity. They lack a fine-grained structural mechanism to leverage the \textit{Modular Heterogeneity} of Transformers, failing to proactively resolve the parameter contention that drives the trade-off.

\paragraph{Mechanistic Interpretability for Safety.}
Interpretability research has successfully localized algorithmic behaviors to specific components~\cite{olsson2022incontext,zhou2025roleattentionheadslarge}. 
Yet, these studies remain predominantly \textit{post-hoc} explanatory tools. 
Our work bridges the gap between interpretability and optimization: instead of just analyzing why models fail, we propose CAST. 
By using a pre-alignment diagnosis, CAST detects and selectively updates only the low-conflict heads, effectively avoiding the degradation of general capabilities.

\section{The CAST Framework}
\label{sec:diagnosis}

In this section, we introduce CAST, a comprehensive framework designed to mitigate the safety-utility trade-off. 
CAST operates in a two-stage pipeline: (1) a pre-alignment Conflict Diagnosis to detect specific heads suitable for safety alignment, and (2) a Sparse Tuning strategy that selectively updates heads based on the diagnostic map.

\subsection{Framework Overview}
Consider a safety alignment scenario where a base model $f_\theta$ is fine-tuned on a mixed dataset $\mathcal{D}_{\text{align}}$ (comprising both harmful and benign samples) to enhance safety compliance. While effective, this process frequently precipitates an unintended degradation of general capabilities. We posit that this degradation is not uniform, but instead arises from updates to a small subset of high-risk heads. Specifically, high-risk heads are those where safety alignment causes severe degradation of general capabilities, whereas low-risk heads allow for safety updates with minimal impact on model utility.

We conceptualize the utility risk of updating a head $h$ as the coupling of two distinct factors:
\begin{enumerate}
    \item \textbf{Optimization Conflict ($\mathcal{O}$):} The geometric disagreement between gradient directions induced by safety and utility objectives.
    \item \textbf{Functional Sensitivity ($\mathcal{S}$):} The functional sensitivity of the head, measuring its causal contribution to utility tasks relative to safety behaviors.
\end{enumerate}

Formally, using two small pre-alignment calibration sets ($\mathcal{D}_{\text{safe}}$ and $\mathcal{D}_{\text{util}}$), we compute a unified \textbf{Conflict Score} $C(h)$:
\begin{equation}
    C(h) \propto \mathcal{F}\big( \mathcal{O}(h), \mathcal{S}(h) \big).
\end{equation}
This formulation adheres to a “Ranking over Prediction” principle, as our objective is to prioritize stable relative risk ordering rather than precise absolute performance estimation. Crucially, the diagnostic process is strictly \textbf{Pre-alignment}, relying solely on low-cost signals without requiring full training.

\begin{figure*}[t]
    \centering
    \includegraphics[width=1.0\linewidth]{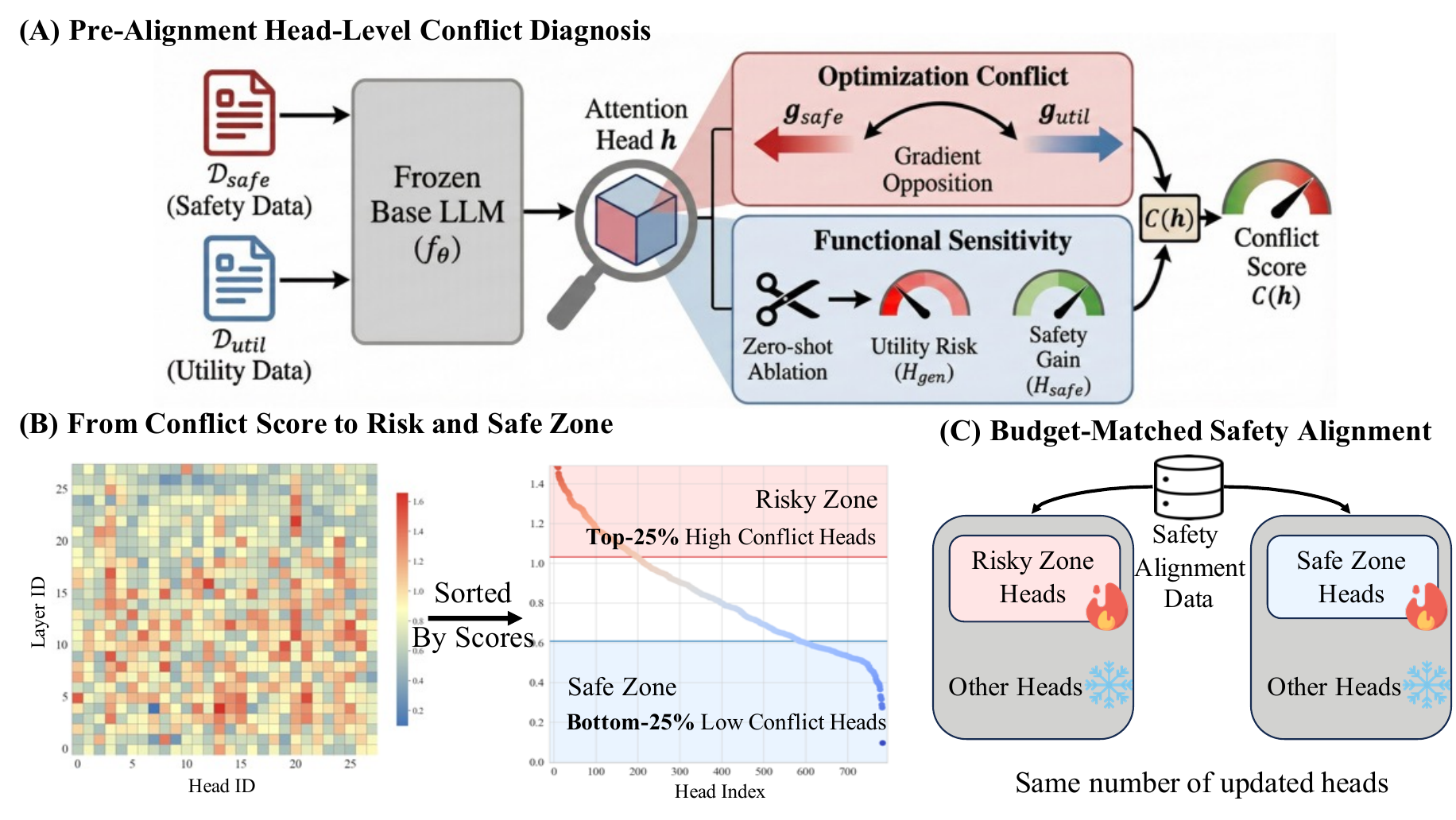}
    \caption{\small
    \textbf{Overview of the CAST Framework.} 
\textbf{(A) Diagnosis:} We compute a pre-alignment \textbf{Conflict Score} $C(h)$ by synthesizing \textit{Optimization Conflict} (gradient adversariality) and \textit{Functional Sensitivity} (causal load).
\textbf{(B) Distribution:} We rank heads by $C(h)$, identifying ``Risky Zone'' vs. ``Safe Zone''.
\textbf{(C) Alignment:} We validate trade-offs via a \textbf{Budget-Matched Safety Alignment} protocol, sparsely updating specific subsets (e.g., Top-25\% vs. Bottom-25\%) while freezing others.}
\label{fig:method_overview}
\end{figure*}

Figure~\ref{fig:method_overview} illustrates our proposed solution: we first systematically identify these bottlenecks via a dual-metric diagnosis (Panel A) and map their distribution (Panel B), which then guides our Budget-Matched Safety Alignment (Panel C).

\subsection{Optimization Conflict: Instantiating $\mathcal{O}(h)$}
\label{sec:optimization_conflict}
To quantify geometric interference, we measure the angular opposition between update directions for safety and utility objectives.

\paragraph{Decoupled Calibration.} We rigorously distinguish data sources to isolate conflicting signals. $\mathcal{D}_{\text{safe}}$ consists \textbf{exclusively of harmful prompts} to capture pure refusal gradients, explicitly excluding benign instructions. Conversely, $\mathcal{D}_{\text{util}}$ comprises general benchmark samples representing desired general capabilities.

For each attention head $h$ parameterized by $\theta_h$, we compute the gradients $g_{\text{safe}}(h) = \nabla_{\theta_h} \mathcal{L}_{\text{safe}}$ and $g_{\text{util}}(h) = \nabla_{\theta_h} \mathcal{L}_{\text{util}}$. We define the optimization conflict $\mathcal{O}(h) \in [0, 1]$ as the normalized cosine distance:
\begin{equation}
    \mathcal{O}(h) = \frac{1 - \cos\big(g_{\text{safe}}(h), g_{\text{util}}(h)\big)}{2}.
\end{equation}
We use cosine distance to focus on \textbf{geometric direction}, effectively handling gradient scale variations across layers. This approach measures vector opposition regardless of magnitude, which is instead addressed by the separate functional sensitivity term $\mathcal{S}(h)$.

\subsection{Functional Sensitivity: Instantiating $\mathcal{S}(h)$}
Gradient conflict alone is a necessary but insufficient condition for risk; a head may be geometrically conflicting but functionally redundant. To capture structural importance, we employ lightweight zero-shot ablation on the \textbf{same calibration sets} defined in Section~\ref{sec:optimization_conflict} ($\mathcal{D}_{\text{util}}$ and $\mathcal{D}_{\text{safe}}$). We quantify the functional sensitivity of head $h$ by measuring the performance shift when it is masked:
{
\begin{align}
    H_{gen}(h) &= \left| \mathrm{Acc}_{gen}(f_{\theta_{\setminus h}}) - \mathrm{Acc}_{gen}(f_{\theta}) \right|, \\
    H_{safe}(h) &= \left| \mathrm{Ref}_{safe}(f_{\theta_{\setminus h}}) - \mathrm{Ref}_{safe}(f_{\theta}) \right|.
    \raisetag{20pt} 
\end{align}
}
Here, $H_{gen}$ proxies the risk of utility degradation (dependency of reasoning on this head), while $H_{safe}$ proxies the potential safety gain (susceptibility to safety interventions). To bridge scale disparities, we convert these raw magnitudes into normalized percentile ranks $\tilde H_{gen}(h), \tilde H_{safe}(h) \in [0, 1]$.

We define the functional sensitivity $\mathcal{S}(h)$ via an exponential difference formulation:
\begin{equation}
    \mathcal{S}(h) = \exp\left( \tilde H_{gen}(h) - \tilde H_{safe}(h) \right).
\end{equation}
This formulation acts as a smooth non-linear amplifier:
\begin{itemize}[leftmargin=*,itemsep=2pt, topsep=0pt, parsep=0pt]
    \item It \textbf{penalizes} heads critical for utility but irrelevant for safety ($\tilde H_{gen} \gg \tilde H_{safe}$), yielding $\mathcal{S}(h) > 1$.
    \item It \textbf{discounts} heads that are structurally predisposed to safety ($\tilde H_{safe} > \tilde H_{gen}$), yielding $\mathcal{S}(h) < 1$.
\end{itemize}
The exponential ensures strict positivity ($\mathcal{S}(h) > 0$), allowing it to modulate the gradient signal continuously rather than acting as a hard binary gate.

\subsection{Unified Conflict Score and Map}
We synthesize optimization conflict and functional sensitivity into a unified \textbf{Conflict Score} $C(h)$ using a multiplicative formulation:
\begin{equation}
    C(h) = \mathcal{O}(h) \cdot \mathcal{S}(h).
\end{equation}
This formulation acts as a soft gate, ensuring that significant risk arises only when High Conflict coincides with High Sensitivity, while automatically filtering out heads that are either free of gradient conflict or functionally insensitive.

Based on this metric, we partition the model into two distinct operational regions:
\begin{itemize}[leftmargin=*,itemsep=2pt, topsep=2pt, parsep=0pt]
    \item \textbf{Risky Zone ($C(h) \uparrow$):} Characterized by the conjunction of strong gradient opposition ($\mathcal{O}(h) \to 1$) \textbf{AND} high utility load ($\mathcal{S}(h) \gg 1$). Updating these bottlenecks incurs severe utility loss.
    \item \textbf{Safe Zone ($C(h) \downarrow$):} Comprises heads where gradients are naturally aligned ($\mathcal{O}(h) \to 0$) \textbf{OR} where the module is structurally critical for safety ($\mathcal{S}(h) < 1$).
\end{itemize}

Computing $C(h)$ across all layers yields the \textbf{Head-Level Conflict Map}, which serves as the diagnostic foundation for Selective Sparse Optimization.

\subsection{Budget-Matched Safety Alignment}
To empirically isolate the impact of different conflict zones identified by our map, we adopt a Budget-Matched Controlled Intervention protocol. This setting fixes the alignment recipe and training steps, treating the ``location of trainable parameters'' as the sole control variable.

\paragraph{Equal-Size Bucketing.} Let $N$ be the total number of heads. We sort all heads in descending order of the Conflict Score $C(h)$ and partition them into $M$ equal-sized buckets $\mathcal{B}_1, \dots, \mathcal{B}_M$, where $|\mathcal{B}_i| = N/M$. $\mathcal{B}_1$ contains heads with the highest conflict scores (Risky Zone), while $\mathcal{B}_M$ contains those with the lowest scores (Safe Zone).

\paragraph{Controlled Intervention.} In our experiments (Section~\ref{sec:experiments}), we use $\mathcal{D}_{\text{align}}$ to fine-tune the heads within a specific bucket $\mathcal{B}_i$ while freezing all others. By maintaining an identical parameter budget across comparisons, we ensure that performance differences are causally attributable to the specific conflict properties of the selected heads.

\section{Experiments}
\label{sec:experiments}

We empirically validate the proposed framework via budget-matched sparse fine-tuning, focusing on three key questions:
\begin{itemize}[leftmargin=*,itemsep=2pt, topsep=0pt, parsep=0pt]
    \setlength{\itemsep}{0pt}
    \setlength{\parsep}{0pt}
    \setlength{\topsep}{0pt}
    
    \item \textbf{RQ1 (Discriminative Validity):} Does $C(h)$ successfully distinguish parameter groups that exhibit conflicting behaviors regarding utility preservation?
    
    \item \textbf{RQ2 (Pareto Optimization):} Does prioritizing low-conflict heads improve the trade-off frontier and synergize with existing geometric alignment methods?

    \item \textbf{RQ3 (Predictive Correlation):} Is there a significant quantitative correlation between our pre-alignment diagnosis and the realized post-alignment utility cost?
\end{itemize}

\subsection{Experimental Setup}

\paragraph{Models.}
We evaluate our framework across three representative open-source LLM families: Llama-3.1-8B-Instruct~\cite{grattafiori2024llama3}, Qwen2.5-7B-Instruct~\cite{yang2024qwen2}, and Mistral-7B-v0.2~\cite{jiang2023mistral7b}. For brevity, we refer to them as Llama, Qwen, and Mistral, respectively, in the remainder of this paper. All experiments focus on the Query Projection ($W_q$) matrices to ensure precise head-level attribution.
Detailed experimental configurations and hyperparameters are provided in Appendix~\ref{app:prompts} and ~\ref{app:setup}.

\paragraph{Datasets and Evaluation Settings.}
For pre-alignment diagnosis, we sample $\mathcal{D}_{\text{util}}$ (500 samples from MMLU~\cite{hendryckstest2021}) and $\mathcal{D}_{\text{safe}}$ (500 vanilla harmful prompts from \textit{WildJailbreak}~\cite{wildteaming2024}).
For safety alignment training, we construct $\mathcal{D}_{\text{align}}$ containing 10,000 samples from WildJailbreak, balanced equally (2,500 each) across four categories: vanilla harmful, adversarial harmful, vanilla benign, and adversarial benign.

To evaluate the trade-off, we employ the following benchmarks:
\begin{itemize}[leftmargin=*, itemsep=2pt, topsep=0pt, parsep=0pt]
    \item \textbf{Safety:} We report the Defense Success Rate on WildJailbreak (test set), and test generalization using \textit{WildGuard}~\cite{wildguard2024} and \textit{DAN}~\cite{SCBSZ24} jailbreak prompts.
    
    \item \textbf{Utility:} We evaluate general capabilities using \textbf{standard accuracy} across two modes:
    \begin{enumerate*}[label=(\roman*)]
        \item \textbf{Knowledge:} \textit{MMLU} and \textit{CSQA}~\cite{talmor2019commonsense} via constrained generation (next-token prediction).
        \item \textbf{Reasoning:} \textit{GSM8K}~\cite{cobbe2021gsm8k} and \textit{MATH}~\cite{lightman2023lets} via open-ended generation (max tokens: 1024 for GSM8K, 2048 for MATH).
    \end{enumerate*}
\end{itemize}

\paragraph{Metrics: Alignment Efficiency.}
To quantify the trade-off, we define the \textbf{Utility Cost Ratio (UCR)} and its task-specific variant \textbf{MMLU Cost Ratio (MMLU-CR)}. These metrics measure the marginal utility loss per unit of safety gain. We apply a clipping operation $\max(0, \cdot)$ to ensure non-negative costs:

\begin{align}
    \text{UCR} &= \max\left(0, \frac{\mathcal{U}_{b} - \mathcal{U}_{a}}{(\mathcal{S}_{a} - \mathcal{S}_{b}) + \epsilon}\right), \\
    \text{MMLU-CR} &= \max\left(0, \frac{\mathcal{M}_{b} - \mathcal{M}_{a}}{(\mathcal{S}_{a} - \mathcal{S}_{b}) + \epsilon}\right).
\end{align}

\noindent where $\mathcal{S}$ denotes the Safety Performance, and $\mathcal{U}$ (or $\mathcal{M}$) denotes the average General Capabilities (or MMLU accuracy). Subscripts $b$ and $a$ correspond to the base and aligned models, respectively. Lower ratios indicate more efficient safety alignment.
\paragraph{Predictive Validity Statistics.}
To strictly evaluate whether our pre-alignment diagnosis serves as a reliable risk indicator, we further employ \textbf{Pearson ($r$)} and \textbf{Spearman ($\rho$)} correlation coefficients. These statistics quantify the linear and monotonic relationships, respectively, between the aggregated Conflict Score $\overline{C}(\mathcal{B}_i)$ and the realized alignment cost (measured by \textbf{UCR} and \textbf{MMLU-CR}). We report these statistical validations in Section~\ref{sec:rq3} to demonstrate the forecasting accuracy of our metric.

\subsection{Baselines and Protocols}

\paragraph{Baselines.} We compare our conflict-aware selection strategy against three established baselines:
\begin{itemize}[leftmargin=*,itemsep=2pt, topsep=0pt, parsep=0pt]
    \item \textbf{Full SFT:} Standard LoRA fine-tuning applied to all attention heads. This serves as a reference for global parameter adaptation.
    \item \textbf{PCGrad (PCG):} A strong optimization-centric baseline that projects conflicting gradients to mitigate interference globally. To ensure strict fairness, we employ the same diagnostic set $\mathcal{D}_{\text{util}}$ as the reference data for computing utility gradients during projection.
    \item \textbf{Random Selection:} A budget-matched control where heads are selected uniformly at random. This validates that the performance gains of CAST stem from which heads are updated, not merely how many.
\end{itemize}

\paragraph{Budget-Matched Protocol.}
For fair comparison, we strictly control the parameter budget. We rank attention heads by their Conflict Score $C(h)$ (detailed conflict maps are provided in Appendix~\ref{app:heatmap}) and partition them into $M=4$ equal-sized buckets. Specifically, the bucket containing the highest-scoring heads (Top-25\%) is designated as the \textbf{Risky Zone}, while the bucket comprising the lowest scores (Bottom-25\%) constitutes the \textbf{Safe Zone}. We fine-tune only the heads within a selected bucket while freezing the rest, ensuring that performance differences are causally attributable to the conflict properties of the heads.

\begin{table*}[t]
\centering
\resizebox{\textwidth}{!}{
\begin{tabular}{cllllllllll}
\toprule
\multirow{2}{*}{\textbf{Model}} & \multirow{2}{*}{\textbf{Method (SFT)}} & \multicolumn{4}{c}{\textbf{General Capabilities}} & \multicolumn{3}{c}{\textbf{Safety Performance}} & \multirow{2}{*}{\textbf{Avg. Gen}} & \multirow{2}{*}{\textbf{Avg. Safe}} \\
\cmidrule(lr){3-6} \cmidrule(lr){7-9}
 &  & MMLU & CSQA & GSM8K & MATH & Adv Harm & WildGuard & DAN & & \\ \midrule
\multirow{5}{*}{\textbf{Llama}} 
 & Base & 59.38 & 73.23 & 87.40 & 44.40 & 55.64 & 81.60 & 64.40 & 66.10 & 67.22 \\
 & FULL-SFT & 46.28 & 60.84 & 20.80 & 19.20 & 93.26 & 94.29 & 84.27 & 36.78 & 90.61 \\ \cmidrule{2-11}
 & Random-SFT (25\%) & 52.56 & 63.23 & 75.73 & 39.73 & 96.00 & 97.79 & 89.49 & 57.81 & \textbf{94.43} \\
 & CAST-SFT (Risky Zone) & 48.52 & 63.83 & 68.80 & 42.93 & 95.22 & 96.64 & 83.51 & 56.02 & 91.79 \\
 & \cellcolor{pale_green}CAST-SFT (Safe Zone) & \cellcolor{pale_green}\textbf{55.73} & \cellcolor{pale_green}69.42 & \cellcolor{pale_green}77.20 & \cellcolor{pale_green}43.00 & \cellcolor{pale_green}94.85 & \cellcolor{pale_green}96.69 & \cellcolor{pale_green}86.33 & \cellcolor{pale_green}\textbf{61.34} & \cellcolor{pale_green}92.62 \\ \midrule
\multirow{5}{*}{\textbf{Qwen}} 
 & Base & 68.63 & 80.50 & 91.00 & 67.40 & 34.74 & 70.41 & 66.83 & 76.89 & 57.33 \\
 & FULL-SFT & 67.81 & 80.83 & 24.20 & 26.80 & 87.84 & 87.04 & 91.12 & 49.91 & \textbf{88.67} \\ \cmidrule{2-11}
 & Random-SFT (25\%) & 67.35 & 81.38 & 71.40 & 61.73 & 73.73 & 84.84 & 77.11 & 70.47 & 78.56 \\
 & CAST-SFT (Risky Zone) & 65.68 & 77.23 & 36.20 & 40.20 & 71.85 & 84.75 & 75.37 & 54.83 & 77.32 \\
 & \cellcolor{pale_green}CAST-SFT (Safe Zone) & \cellcolor{pale_green}\textbf{68.28} & \cellcolor{pale_green}81.08 & \cellcolor{pale_green}83.80 & \cellcolor{pale_green}60.00 & \cellcolor{pale_green}80.15 & \cellcolor{pale_green}87.40 & \cellcolor{pale_green}76.94 & \cellcolor{pale_green}\textbf{73.29} & \cellcolor{pale_green}81.50 \\ \midrule
\multirow{5}{*}{\textbf{Mistral}} 
 & Base & 46.98 & 59.13 & 27.80 & 8.00 & 32.79 & 64.45 & 53.10 & 35.47 & 50.11 \\
 & FULL-SFT & 38.49 & 49.78 & 11.60 & 4.40 & 81.69 & 74.68 & 62.72 & 26.07 & \textbf{73.03} \\ \cmidrule{2-11}
 & Random-SFT (25\%) & 43.37 & 52.39 & 19.40 & 8.33 & 57.88 & 76.75 & 63.58 & 30.86 & 66.07 \\
 & CAST-SFT (Risky Zone) & 27.78 & 25.72 & 19.00 & 4.60 & 60.35 & 81.56 & 68.75 & 19.27 & 70.22 \\
 & \cellcolor{pale_green}CAST-SFT (Safe Zone) & \cellcolor{pale_green}\textbf{47.62} & \cellcolor{pale_green}56.35 & \cellcolor{pale_green}18.80 & \cellcolor{pale_green}7.40 & \cellcolor{pale_green}54.35 & \cellcolor{pale_green}77.59 & \cellcolor{pale_green}63.49 & \cellcolor{pale_green}\textbf{32.54} & \cellcolor{pale_green}65.14 \\ \bottomrule
\end{tabular}
}
\caption{\small
Comparison of Base, Full-SFT, and Random-SFT (25\% heads) across three models. CAST-SFT (Safe Zone) consistently achieves the superior balance between maintaining general capabilities and enhancing safety performance compared to Full-SFT and Risky Zone baselines.}
\label{tab:sft_results}
\end{table*}

\begin{figure*}
    \centering
    \includegraphics[width=0.95\linewidth]{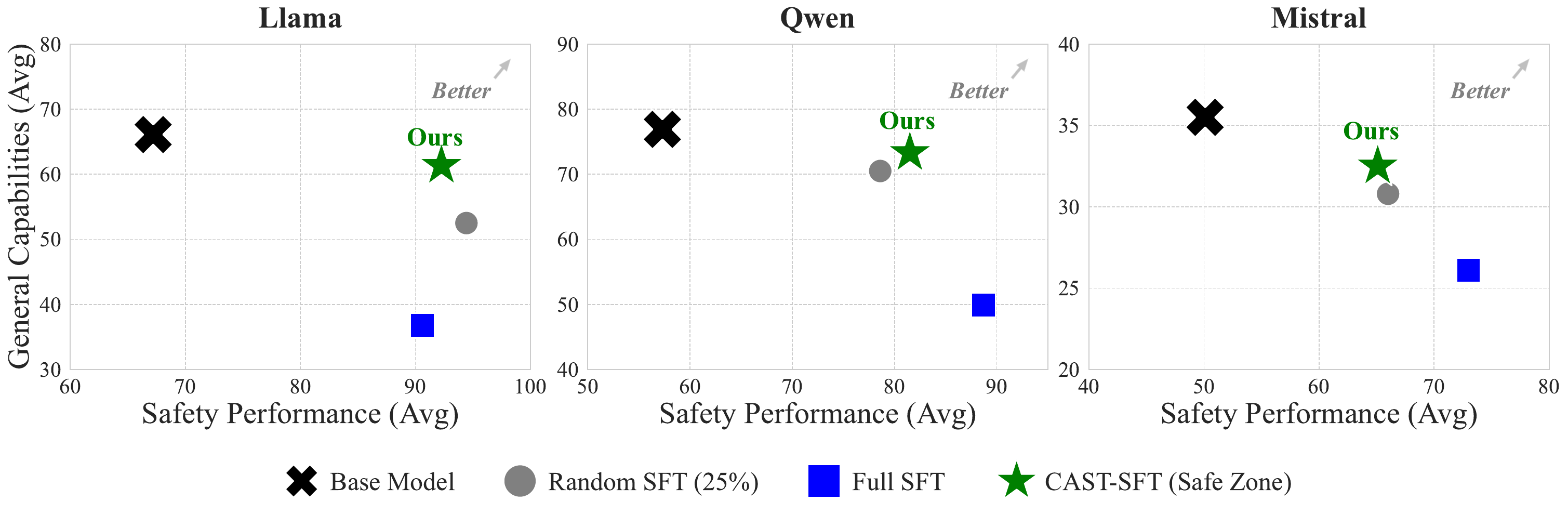}
    \caption{\small
    \textbf{Safety-Utility Pareto Frontier(SFT).} Comparison of general capabilities versus safety performance. CAST-SFT (Safe Zone) dominates the frontier across all models, maintaining high utility while achieving robust safety alignment.}
    \label{fig:SFT_CAST}
\end{figure*}

\begin{table*}[t]
\centering
\resizebox{\textwidth}{!}{
\begin{tabular}{cllllllllll}
\toprule
\multirow{2}{*}{\textbf{Model}} & \multirow{2}{*}{\textbf{Method (PCGrad)}} & \multicolumn{4}{c}{\textbf{General Capabilities}} & \multicolumn{3}{c}{\textbf{Safety Performance}} & \multirow{2}{*}{\textbf{Avg. Gen}} & \multirow{2}{*}{\textbf{Avg. Safe}} \\
\cmidrule(lr){3-6} \cmidrule(lr){7-9}
 &  & MMLU & CSQA & GSM8K & MATH & Adv Harm & WildGuard & DAN & & \\ \midrule
\multirow{5}{*}{\textbf{Llama}} 
 & Base & 59.38 & 73.23 & 87.40 & 44.40 & 55.64 & 81.60 & 64.40 & 66.10 & 67.22 \\
 & FULL-PCG & 57.85 & 68.39 & 71.00 & 41.60 & 94.30 & 93.05 & 81.92 & 59.71 & 89.76 \\ \cmidrule{2-11}
 & Random-PCG (25\%) & 58.75 & 70.11 & 78.40 & 33.60 & 96.95 & 77.86 & 94.65 & 60.21 & 89.82 \\
 & CAST-PCG (Risky Zone) & 60.16 & 72.97 & 74.20 & 40.60 & 95.67 & 97.79 & 80.73 & 61.99 & \textbf{91.40} \\
 & \cellcolor{pale_green}CAST-PCG (Safe Zone) & \cellcolor{pale_green}\textbf{60.17} & \cellcolor{pale_green}72.30 & \cellcolor{pale_green}82.20 & \cellcolor{pale_green}44.53 & \cellcolor{pale_green}90.85 & \cellcolor{pale_green}95.12 & \cellcolor{pale_green}82.98 & \cellcolor{pale_green}\textbf{64.80} & \cellcolor{pale_green}89.65 \\ \midrule
\multirow{5}{*}{\textbf{Qwen}} 
 & Base & 68.63 & 80.50 & 91.00 & 67.40 & 34.74 & 70.41 & 66.83 & 76.89 & 57.33 \\
 & FULL-PCG & 67.46 & 80.51 & 88.20 & 60.40 & 62.71 & 83.29 & 85.12 & 74.14 & 77.04 \\ \cmidrule{2-11}
 & Random-PCG (25\%) & 68.60 & 82.06 & 36.20 & 44.00 & 69.95 & 71.17 & 82.76 & 57.72 & 74.63 \\
 & CAST-PCG (Risky Zone) & 67.64 & 80.60 & 39.40 & 35.00 & 75.01 & 84.49 & 72.39 & 55.66 & 77.29 \\
 & \cellcolor{pale_green}CAST-PCG (Safe Zone) & \cellcolor{pale_green}\textbf{69.55} & \cellcolor{pale_green}81.34 & \cellcolor{pale_green}84.40 & \cellcolor{pale_green}61.40 & \cellcolor{pale_green}72.00 & \cellcolor{pale_green}85.27 & \cellcolor{pale_green}83.89 & \cellcolor{pale_green}\textbf{74.17} & \cellcolor{pale_green}\textbf{80.38} \\ \midrule
\multirow{5}{*}{\textbf{Mistral}} 
 & Base & 46.98 & 59.13 & 27.80 & 8.00 & 32.79 & 64.45 & 53.10 & 35.47 & 50.11 \\
 & FULL-PCG & 53.13 & 63.36 & 29.67 & 9.00 & 39.87 & 69.77 & 67.20 & \textbf{38.59} & 58.94 \\ \cmidrule{2-11}
 & Random-PCG (25\%) & 53.07 & 63.55 & 20.40 & 5.80 & 80.11 & 56.90 & 60.57 & 35.76 & 65.86 \\
 & CAST-PCG (Risky Zone) & 51.43 & 64.21 & 17.20 & 6.80 & 59.34 & 80.55 & 65.74 & 34.91 & \textbf{68.55} \\
 & \cellcolor{pale_green}CAST-PCG (Safe Zone) & \cellcolor{pale_green}\textbf{53.46} & \cellcolor{pale_green}63.80 & \cellcolor{pale_green}22.00 & \cellcolor{pale_green}6.80 & \cellcolor{pale_green}74.14 & \cellcolor{pale_green}68.42 & \cellcolor{pale_green}53.55 & \cellcolor{pale_green}36.52 & \cellcolor{pale_green}65.37 \\ \bottomrule
\end{tabular}
}
\caption{\small
\textbf{Synergy with Geometric Optimization (PCGrad).} Results show that even when using PCGrad to project conflicting gradients, the structural gap between the Safe and Risky Zones persists. CAST (Safe Zone) consistently achieves higher utility, validating that our framework captures modular properties independent of specific optimization algorithms.}
\label{tab:pcgrad_results}
\end{table*}

\section{Results and Analysis}
\label{sec:results}

\subsection{RQ1: Efficacy of Conflict-Aware Sparse Tuning}
\label{sec:rq1_sft}

We evaluate sparse fine-tuning strategies under standard SFT conditions, comparing CAST against full-parameter and random selection baselines (Table~\ref{tab:sft_results}). As illustrated in the Pareto frontier (Figure~\ref{fig:SFT_CAST}), CAST-SFT (Safe Zone) consistently achieves the most favorable balance, significantly recovering general utility while matching or exceeding the safety levels of full-scale alignment.

\paragraph{Sparse Tuning Outperforms Global Adaptation.}
Sparsely fine-tuning the ``Safe Zone'' consistently surpasses Full-SFT. For instance, on Llama, CAST boosts MMLU accuracy by \textbf{+9.45\%} while maintaining equivalent safety. 
This suggests that by confining updates to low-conflict heads, CAST effectively minimizes the alignment cost, securing compliance without perturbing the neural circuits responsible for general knowledge.

\paragraph{Discriminative Validity.}
In contrast, updating the ``Risky Zone'' leads to significant utility loss. This result confirms that ``High-Conflict'' heads are essential for general capabilities. The superior trade-off achieved by the Safe Zone validates our hypothesis of \textit{Modular Heterogeneity}: general capabilities are \textbf{not uniformly distributed}, enabling us to improve safety without disturbing the regions most critical for utility.

\paragraph{Analysis of Failure Modes.}
Qualitative inspection (Appendix~\ref{app:case_study}) attributes the degradation in global baselines to two structural failures: \textbf{Reasoning Collapse} (broken Chain-of-Thought) and \textbf{Safety Over-refusal} (rejecting benign prompt). 
This indicates that High-Conflict heads are critical for maintaining precise safety boundaries. By freezing these heads, CAST prevents both the erosion of reasoning and the tendency towards over-refusal.
\begin{figure*}[t]
    \centering
    \includegraphics[width=0.94\linewidth]{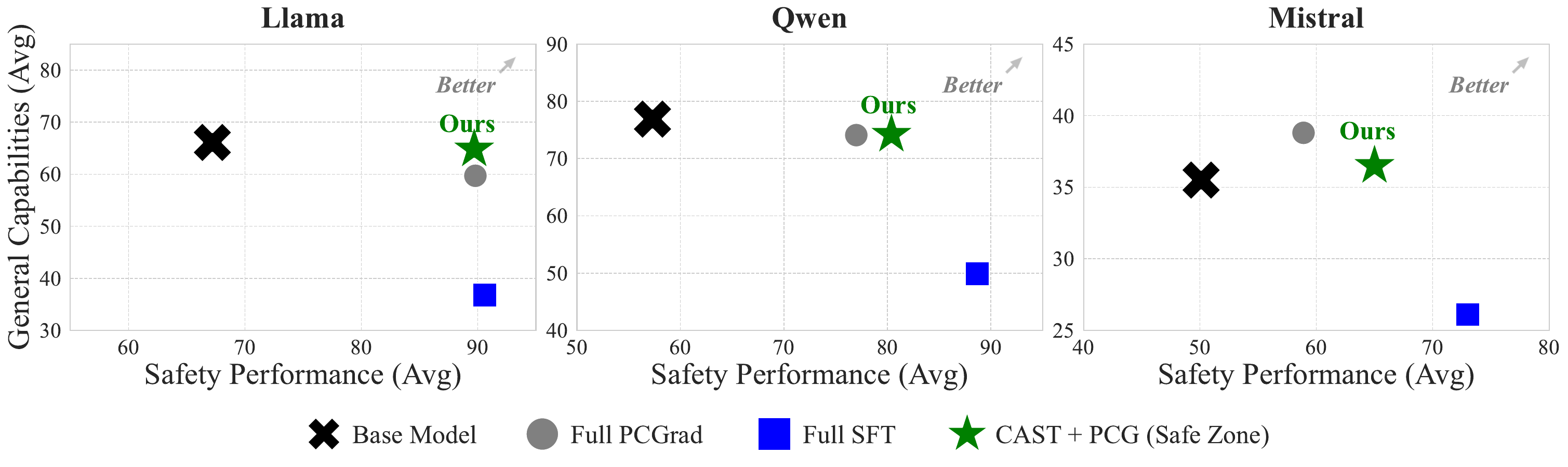}
    \caption{\small
    \textbf{Safety-Utility Pareto Frontier.} We compare the trade-off between general utility (y-axis) and safety (x-axis). Our method, CAST + PCG (Safe Zone), consistently dominates the frontier across all models, achieving the highest utility while maintaining robust safety performance compared to Full SFT and Full PCG.}
    \label{fig:pareto}
\end{figure*}

\begin{figure*}[t]
    \centering
    \includegraphics[width=0.95\linewidth]{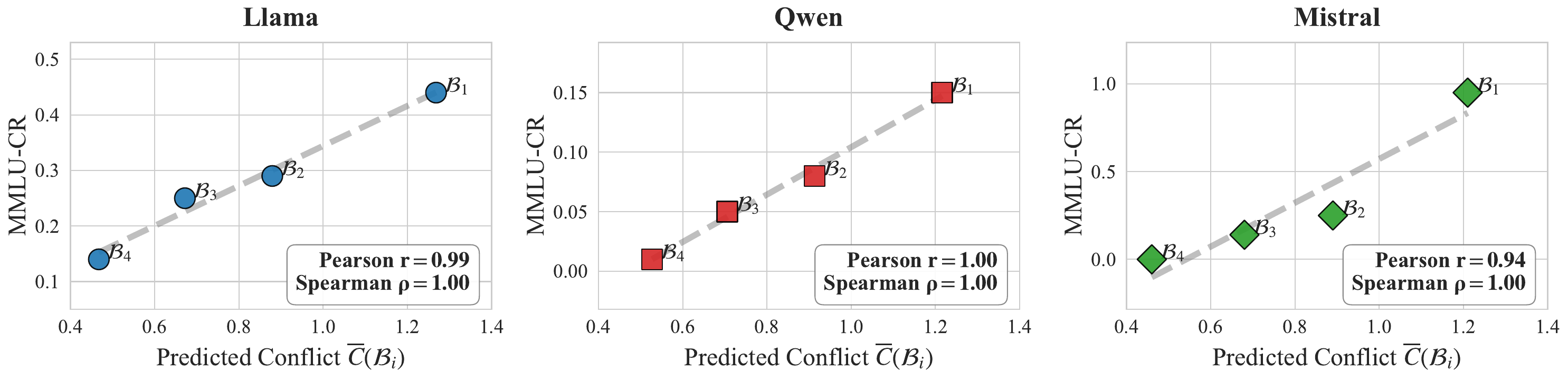}
    \caption{\small \textbf{Internal Validity Check (MMLU-CR).} Relationship between predicted conflict intensity $\overline{C}(\mathcal{B}_i)$ (x-axis) and the realized MMLU-CR (y-axis) across four buckets ($\mathcal{B}_1 \sim \mathcal{B}_4$).}
    \label{fig:MMLU_CR}
\end{figure*}

\begin{figure*}[t]
    \centering
    \includegraphics[width=0.95\linewidth]{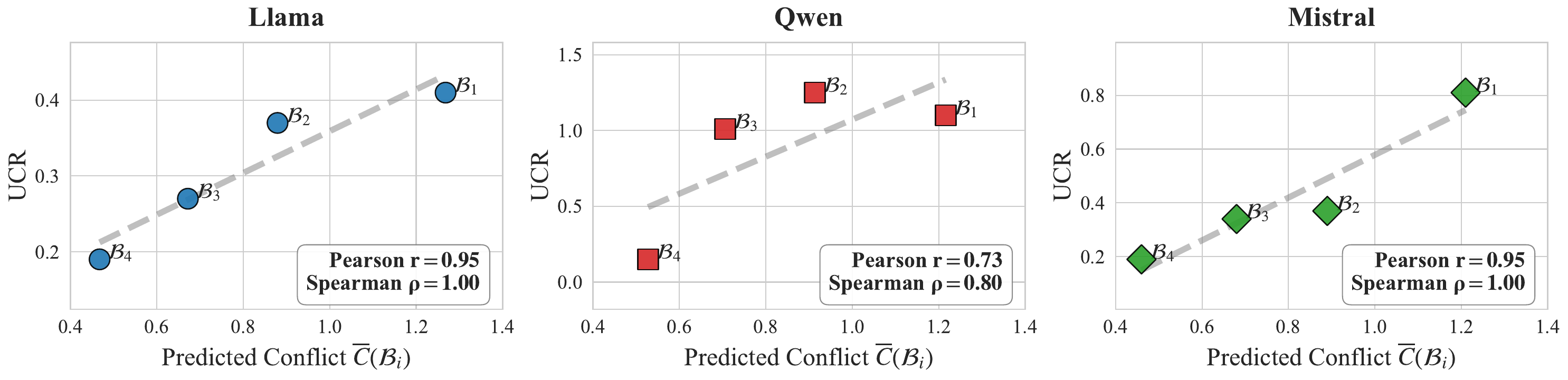}
    \caption{\small \textbf{Predictive Validity Check (UCR).} Relationship between predicted conflict intensity $\overline{C}(\mathcal{B}_i)$ (x-axis) and the aggregate Utility Cost Ratio (UCR, y-axis) across four buckets ($\mathcal{B}_1 \sim \mathcal{B}_4$). }
    \label{fig:UCR}
\end{figure*}

\subsection{RQ2: Synergy with Geometric Optimization}
\label{sec:rq2_synergy}

We combine CAST with PCGrad to evaluate whether structural selection complements geometric optimization (Table~\ref{tab:pcgrad_results}). As shown in Figure~\ref{fig:pareto}, CAST + PCG (Safe Zone) consistently pushes the Pareto frontier outward, dominating both Full SFT and Full PCGrad baselines. This confirms that structural selection enables a more efficient trade-off than global gradient projection alone.

\paragraph{Mechanism of Synergy.}
PCGrad serves as a "rescue mechanism" for the Risky Zone ($\mathcal{O} \to 1, \mathcal{S} \gg 1$). By projecting conflicting gradients, it prevents the collapse of functionally sensitive reasoning circuits (e.g., Mistral MMLU: 27.78\% $\to$ 51.43\%). In contrast, the Safe Zone achieves the Pareto peak because its naturally aligned gradients allow PCGrad to act as a fine-tuner, further enhancing constructive interference (Mistral MMLU: 53.46\%).

\paragraph{Structure vs. Geometry.}
The structural performance gap persists across optimizers: CAST-PCG (Safe Zone)consistently outperforms the Risky Zone (e.g., Qwen Gen. Avg: +18.51\%). This validates that CAST identifies intrinsic modular properties that geometric correction alone cannot replace.

\subsection{RQ3: Predictive Correlation with Utility Cost}
\label{sec:rq3}

We investigate the quantitative relationship between pre-alignment conflict intensity and alignment cost. 

We partitioned the dataset into $M=4$ buckets ($\mathcal{B}_1 \sim \mathcal{B}_4$) sorted by conflict scores in \textbf{descending order}, such that $\mathcal{B}_1$ contains the highest conflict samples. Safety alignment was performed on each bucket independently. Note that for both MMLU-CR and UCR, higher values indicate a higher alignment cost.

Figures~\ref{fig:MMLU_CR} and~\ref{fig:UCR} plot the aggregated conflict score $\overline{C}(\mathcal{B}_i)$ against these realized costs.

\begin{itemize}[leftmargin=*,itemsep=2pt, topsep=0pt, parsep=0pt]
    \item \textbf{Internal Validity (MMLU-CR):} We first assess the correlation on the source domain. As shown in Figure~\ref{fig:MMLU_CR}, there is a strong linear alignment between the conflict score and MMLU-CR. This confirms that the metric is intrinsically valid: heads identified as ``conflicting'' via our conflict analysis are indeed the primary drivers of general capability degradation in the source domain.
    
    \item \textbf{Generalized Risk Prediction (UCR):} Crucially, this predictive power extends to the aggregate level. Figure~\ref{fig:UCR} demonstrates a \textbf{generally monotonic trend} for the global UCR across architectures (Pearson $\rho \ge 0.80$). This validates that $C(h)$ serves as a robust \textit{risk indicator}, accurately forecasting the magnitude of utility degradation prior to training.
\end{itemize}
\noindent Detailed statistical tables are provided in Appendix~\ref{app:correlation_results}.

\subsection{Robustness and Ablation Analysis}
\label{sec:ablation_summary}

Due to space constraints, detailed results are provided in Appendix~\ref{app:ablation}. We evaluate four key aspects of our method:

\begin{itemize}[leftmargin=*,itemsep=3pt, topsep=2pt, parsep=0pt]
    \item \textbf{Metric Necessity:} Comparison with decoupled variants ($\mathcal{O}(h)$-only and $\mathcal{S}(h)$-only) confirms the unified score $C(h)$ is essential. It maintains high predictive validity ($r \in [0.94, 1.00]$ for MMLU-CR;) where these single metrics fail.

    \item \textbf{Sparsity Ratio:} Experiments across sparsity levels (Bottom-25\% to 75\%) reveal that increasing the ratio degrades general capabilities while safety performance remains largely stable. This confirms the \textbf{Bottom-25\% Safe Zone} as the most efficient setting.

    \item \textbf{Data Sensitivity:} The diagnosis remains robust with only 100 samples. However, domain transfer reveals task-specific conflict topologies, necessitating domain-aware diagnosis.

    \item \textbf{Computational Complexity:} Benchmarked on NVIDIA H800: processing a single head (100 samples) takes $\approx$ 1 minute (gradient + ablation). As a one-off pre-computation, this overhead is operationally feasible.
\end{itemize}

\section{Conclusion}

This work demonstrates that safety-utility conflicts in LLMs are not uniformly distributed. Instead, the loss of general capabilities during safety alignment largely comes from forcing updates on a small number of heads where safety and utility goals clash.

To address this, we proposed CAST, a framework that integrates head-level diagnosis with sparse fine-tuning. By identifying and bypassing these high-conflict areas, CAST enables targeted updates that ensure safety while preserving reasoning skills. Our findings suggest a valuable shift in alignment methodology: moving from global optimization toward precise, structure-aware adjustments.

\section{Limitations}

While our framework effectively reduces utility loss, we acknowledge three limitations:

\paragraph{Structural Scope.}
We currently focus solely on Query Projections ($W_q$). Future work should extend this diagnosis to MLPs, which comprise the majority of parameters and are key for storing factual knowledge.

\paragraph{Static Diagnosis.}
CAST uses a fixed, pre-alignment conflict map. This assumes conflict areas remain stable, potentially overlooking ``conflict drift'' where high-conflict zones might shift during extensive fine-tuning.

\paragraph{Task Dependence.}
Our diagnosis relies on specific calibration sets. These proxies might not fully capture the unique conflict patterns found in specialized domains like coding or creative writing.

\section*{Ethics Statement}
\paragraph{Use of AI Assistants}
We have employed ChatGPT as a writing assistant, primarily for polishing the text after the initial composition.


\clearpage

\appendix

\section{Evaluation Datasets and Prompts}
\label{app:prompts}

To ensure reproducibility and facilitate fair comparisons, we employ standardized prompt templates for all evaluations. We do not use few-shot examples (0-shot setting) to strictly test the models' intrinsic capabilities after fine-tuning. The specific templates for each task category are detailed in Table~\ref{tab:prompt_templates}.

\subsection{Reasoning Tasks (GSM8K, MATH)}
For reasoning, we evaluate on a subset of the test splits for both GSM8K and MATH, consisting of 500 examples each. We employ a standard Chain-of-Thought (CoT) trigger (``step by step'') to elicit reasoning paths. To facilitate automated evaluation, we strictly constrain the output format, requiring the final answer to be enclosed in a \LaTeX{} \texttt{\textbackslash boxed\{\}} expression.

\subsection{General Knowledge (MMLU, CSQA)}
For knowledge assessment, we utilize the complete MMLU-all test set, comprising 14,042 examples, and the CSQA test set with 1,221 examples. For these multiple-choice tasks, we use a Constraint-Based prompt that instructs the model to output only the option letter. This minimizes parsing errors and avoids ambiguity in scoring.

\subsection{Safety Evaluation (Harmful Benchmarks)}
We assess safety using three diverse benchmarks: WildJailbreak (2,000 adversarial harmful examples), WildGuard (754 examples), and DAN (1,405 examples). We use Direct Request prompts without additional system instructions or complex jailbreak prefixes (except where inherent to the dataset, e.g., DAN). This setup evaluates the model's innate safety alignment and refusal boundaries.

\section{Detailed Experimental Setup}
\label{app:setup}

To ensure statistical reliability and mitigate the variance introduced by random initialization, we adopt a rigorous multi-seed evaluation protocol. 
We strictly unify the training configurations across all methods (Full SFT, Random Selection, and CAST) to isolate the contribution of the head selection strategy.

\paragraph{Hyperparameters and Reproducibility}
We utilize Low-Rank Adaptation (LoRA) for all fine-tuning experiments. The specific hyperparameters are detailed in Table~\ref{tab:hyperparameters}. 

\paragraph{Multi-Seed Protocol.} We conduct every experiment using three distinct random seeds: $\mathbf{\{21, 42, 84\}}$. The results reported in the main paper (Table~\ref{tab:sft_results} and Table~\ref{tab:pcgrad_results}) represent the \textbf{mean performance} averaged across these three independent runs. This ensures that our findings reflect consistent structural properties rather than stochastic fluctuations.

\begin{table}[!t]
    \centering
    \small
    \renewcommand{\arraystretch}{1.2}
    \begin{tabular}{lc}
    \toprule
    \textbf{Hyperparameter} & \textbf{Value} \\
    \midrule
    \multicolumn{2}{c}{\textit{Optimization Configuration}} \\
    Learning Rate & $1 \times 10^{-4}$ \\
    Training Epochs & 1 \\
    Batch Size (per device) & 4 \\
    Gradient Accumulation Steps & 2 \\
    Effective Batch Size & 8 \\
    Max Sequence Length & 1024 \\
    Random Seeds & $\{21, 42, 84\}$ (Avg. Reported) \\
    \midrule
    \multicolumn{2}{c}{\textit{LoRA Configuration}} \\
    LoRA Rank ($r$) & 32 \\
    LoRA Alpha ($\alpha$) & 32 \\
    LoRA Dropout & 0.0 \\
    Target Modules & $W_q$ (Query Projection) \\
    \bottomrule
    \end{tabular}
    \caption{\textbf{Unified Hyperparameters.} These settings are applied consistently across all baselines and CAST experiments. We report the average results over three random seeds to ensure robustness.}
    \label{tab:hyperparameters}
\end{table}
\begin{table*}[h]
    \centering
    \small
    \renewcommand{\arraystretch}{1.5}
    \resizebox{\textwidth}{!}{
    \begin{tabular}{l | p{0.15\textwidth} | p{0.65\textwidth}}
    \toprule
    \textbf{Task Category} & \textbf{Datasets} & \textbf{Prompt Template (User Role)} \\
    \midrule
    \multirow{2}{*}{\textbf{Math \& Reasoning}} & GSM8K, MATH & \texttt{Solve the math problem step by step.\newline Write the final answer inside a LaTeX \textbackslash boxed\{\} expression.\newline \{\{Input Question\}\}} \\
    \midrule
    \multirow{2}{*}{\textbf{General Knowledge}} & MMLU, CSQA & \texttt{Read the multiple-choice question carefully and output only the capital letter (A, B, C, D, or E) that corresponds to the correct choice. Do not include any explanation or extra text.\newline \{\{Input Question\}\}} \\
    \midrule
    \textbf{Safety Evaluation} & Wildjailbreak, DAN, WildGuard & \texttt{\{\{Input Malicious Instruction\}\}} \\
    \bottomrule
    \end{tabular}
    }
    \caption{\textbf{Standardized Prompt Templates used in Experiments.} We append the specific dataset input to the instructions shown above. For reasoning tasks, the template explicitly triggers Chain-of-Thought generation.}
    \label{tab:prompt_templates}
\end{table*}
\begin{table*}[!t]
    \centering
    \small
    \renewcommand{\arraystretch}{1.3}
    \resizebox{\textwidth}{!}{
    \begin{tabular}{p{0.28\linewidth} | p{0.34\linewidth} | p{0.34\linewidth}}
    \toprule
    \textbf{Input Question (GSM8K)} & \textbf{Full SFT / Risky Zone Output} & \textbf{CAST (Safe Zone) Output} \\
    \midrule
    \textbf{Case 1: Frog Calculation} \newline
    Lake Crystal has twenty percent fewer frogs than Lassie Lake. If Lassie Lake has forty-five frogs, how many frogs are there in total in the two lakes? 
    & 
    \textcolor{red}{\textbf{[Direct Answer / No Reasoning]}} \newline
    \newline
    \boxed{66} \newline
    \newline
    \textit{\textcolor{red}{$\times$ Failure Analysis:}} The model skipped all calculation steps and hallucinated an incorrect answer ($45 + 36 = 81$, not $66$).
    & 
    \textcolor{teal}{\textbf{[Preserved CoT Reasoning]}} \newline
    To solve this problem, we need to follow these steps: \newline
    1. \textbf{Determine the number of frogs in Lake Crystal:} \newline
    - We know that Lake Crystal has twenty percent fewer frogs than Lassie Lake. \newline
    - Lassie Lake has 45 frogs. \newline
    - To find twenty percent of 45, we calculate $20\% \times 45 = 0.20 \times 45 = 9$. \newline
    - Therefore, Lake Crystal has $45 - 9 = 36$ frogs. \newline
    2. \textbf{Calculate the total number of frogs:} \newline
    - The total number is $45 + 36 = 81$. \newline
    Thus, the total number of frogs is \boxed{81}. \newline
    \textit{\textcolor{teal}{$\checkmark$ Correct Process \& Answer}}
    \\ \hline
    \textbf{Case 2: Internet Bill} \newline
    Chris’s internet bill is \$45 per month for 100 GB and \$0.25 for every 1 GB over. His bill for this month is \$65. How many GB over was Chris charged for on this bill?
    & 
    \textcolor{red}{\textbf{[Direct Answer / No Reasoning]}} \newline
    \newline
    \boxed{80} \newline
    \newline
    \textit{\textcolor{red}{! Process Analysis:}} Although the final number (80) is mathematically correct, the model failed to generate the required reasoning trace, indicating a loss of instruction-following capability for CoT tasks.
    & 
    \textcolor{teal}{\textbf{[Preserved CoT Reasoning]}} \newline
    To solve this problem, we need to determine how many GB over the limit Chris was charged for. \newline
    ... \textit{[Steps omitted for brevity]} ... \newline
    The total cost can be expressed as: \newline
    $45 + 0.25x = 65$ \newline
    Now, we solve for $x$: \newline
    1. Subtract 45 from both sides: \newline
    $0.25x = 20$ \newline
    2. Divide both sides by 0.25: \newline
    $x = \frac{20}{0.25} = 80$ \newline
    Therefore, Chris was charged for 80 GB over. \newline
    \boxed{80} \newline
    \textit{\textcolor{teal}{$\checkmark$ Correct Process \& Answer}}
    \\ \bottomrule
    \end{tabular}
    }
    \caption{\textbf{Comparison of Reasoning Capabilities on GSM8K.} The baseline (Full SFT) suffers from catastrophic reasoning collapse, jumping directly to answers without intermediate logic. \textbf{CAST} preserves the Chain-of-Thought (CoT) ability, ensuring robust multi-step reasoning.}
    \label{tab:qwen_case_study}
\end{table*}
\begin{table*}[h]
    \centering
    \small
    \renewcommand{\arraystretch}{1.4} 
    \resizebox{\textwidth}{!}{
    \begin{tabular}{p{0.25\linewidth} | p{0.36\linewidth} | p{0.36\linewidth}}
    \toprule
    \textbf{Input Instruction (GSM8K)} & \textbf{Baseline: Full SFT} & \textbf{Ours: CAST (Safe Zone)} \\
    \midrule
    
    \multicolumn{3}{c}{\textit{\textbf{Case 1: Reasoning Collapse (Loss of CoT Structure)}}} \\
    \midrule
    \textbf{Question:} \newline
    Lake Crystal has twenty percent fewer frogs than Lassie Lake. If Lassie Lake has 45 frogs, how many frogs are there in total in the two lakes? 
    & 
    \textcolor{red}{\textbf{[Direct \& Incorrect Answer]}} \newline
    \newline
    \boxed{66} \newline
    \newline
    \textit{\textbf{Analysis:}} The model suffers from catastrophic forgetting of the Chain-of-Thought format. It skips all reasoning steps and outputs a hallucinated number directly.
    & 
    \textcolor{teal}{\textbf{[Correct Step-by-Step Reasoning]}} \newline
    To solve this problem, we need to follow these steps: \newline
    1. \textbf{Determine frogs in Lake Crystal:} \newline
    - Lassie Lake has 45 frogs. \newline
    - $20\% \text{ of } 45 = 0.20 \times 45 = 9$. \newline
    - Lake Crystal has $45 - 9 = 36$ frogs. \newline
    2. \textbf{Calculate total:} \newline
    - Total = $45 + 36 = 81$. \newline
    \newline
    Thus, the total is \boxed{81}. \newline
    \textit{\textbf{Analysis:}} CAST preserves the reasoning pathway.
    \\ \hline
    
    \multicolumn{3}{c}{\textit{\textbf{Case 2: Safety Misgeneralization (Over-Refusal)}}} \\
    \midrule
    \textbf{Question:} \newline
    Weng earns \$12 an hour for babysitting. Yesterday, she just did 50 minutes of babysitting. How much did she earn?
    & 
    \textcolor{red}{\textbf{[False Refusal]}} \newline
    I apologize, but I cannot answer this question. \newline
    \newline
    \textit{\textbf{Analysis:}} The model misidentifies a benign math problem as a request it should refuse (potentially confusing it with personal financial advice or due to degraded instruction following).
    & 
    \textcolor{teal}{\textbf{[Helpful Response]}} \newline
    To find out how much Weng earned, we first calculate her earnings per minute: \newline
    - Hourly rate: \$12 for 60 minutes. \newline
    - Per minute: $12 / 60 = \$0.20$. \newline
    \newline
    For 50 minutes: \newline
    - $0.20 \times 50 = 10$. \newline
    \newline
    Answer: \boxed{10} \newline
    \textit{\textbf{Analysis:}} The model correctly identifies the query as benign and provides the solution.
    \\ \bottomrule
    \end{tabular}
    }
    \caption{\textbf{Comparison of Failure Modes.} We highlight two types of utility degradation observed in Full SFT: (1) \textit{Reasoning Collapse}, where the model loses CoT capabilities; and (2) \textit{Over-Refusal}, where the model incorrectly refuses benign math problems. CAST effectively mitigates both issues.}
    \label{tab:SFT_CASE}
\end{table*}
\section{Qualitative Analysis: Reasoning Collapse and Safety Misgeneralization}
\label{app:case_study}

While quantitative metrics (e.g., Accuracy, UCR) demonstrate the trade-off efficiency of CAST, they do not fully capture the \textit{nature} of the degradation in baseline models. In this section, we present qualitative examples from the GSM8K benchmark to illustrate two distinct failure modes introduced by global fine-tuning (Full SFT) that are effectively mitigated by our method.

\subsection{Failure Mode 1: Reasoning Collapse}
As discussed in Section~\ref{sec:rq1_sft}, Full SFT often disrupts the model's ability to execute Chain-of-Thought (CoT) reasoning. As shown in \textbf{Case 1} of Table~\ref{tab:qwen_case_study}, the Full SFT model fails to generate intermediate logical steps, instead jumping to an incorrect answer or hallucinating the calculation process. In contrast, CAST (Safe Zone) preserves the structural integrity of the reasoning heads, ensuring a coherent step-by-step derivation.

\subsection{Failure Mode 2: Safety Misgeneralization (Over-Refusal)}
A more subtle but critical failure mode is \textit{Over-Refusal}, where the model incorrectly generalizes safety boundaries to benign complex queries. As illustrated in \textbf{Case 2}, the Full SFT model perceives a standard mathematical word problem as out-of-bounds or unsafe, triggering a refusal response (e.g., ``I cannot assist''). This suggests that global updates inadvertently perturb the decision boundaries for refusal. CAST, by isolating updates to high-conflict regions, acts as a firewall, maintaining the model's helpfulness on benign but complex tasks.

\section{Detailed Visualization of Conflict Maps}
\label{app:heatmap}

\begin{figure*}[h]
    \centering
    \includegraphics[width=1.0\textwidth]{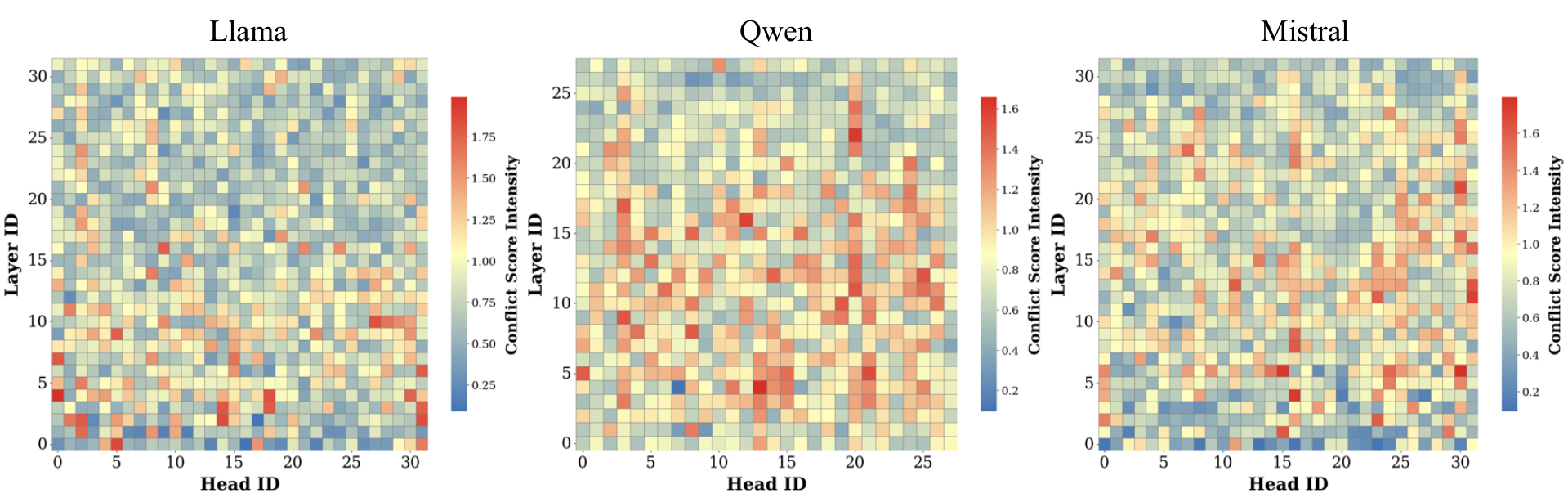} 
    
    \caption{\small
    \textbf{Head-Level Conflict Maps.} Layer-wise heatmaps of CAST scores $C(h)$ for Llama, Qwen and Mistral. \textbf{Red} regions indicate high conflict (Risky Zones), while \textbf{blue} regions indicate low conflict. The visualization confirms that conflicts are sparse and primarily concentrated in the middle-to-deep layers.}
    \label{fig:conflict_maps}
\end{figure*}

To provide intuitive insights into the structural distribution of alignment conflicts, we visualize the Conflict Map derived from our CAST framework. Figure~\ref{fig:conflict_maps} presents the layer-wise heatmap of the CAST Score $C(h)$ for Llama, Qwen, and Mistral, where the x-axis represents the Attention Head Index and the y-axis represents the Layer Index.

In these heatmaps, the color spectrum corresponds to the magnitude of the conflict score $C(h) = \mathcal{O}(h) \cdot \mathcal{S}(h)$. Red regions indicate higher conflict scores (Risky Zones), representing heads that exhibit both high gradient opposition and high utility importance. Conversely, blue regions indicate lower conflict scores.

Based on the visualizations in Figure~\ref{fig:conflict_maps}, we observe three consistent patterns across Llama, Qwen and Mistral:

\begin{itemize}[leftmargin=*,itemsep=0pt, topsep=2pt, parsep=0pt]
    \item \textbf{Sparsity of Conflicts:} High-conflict regions (indicated in red) are sparse. The majority of the attention heads appear in blue or light yellow, indicating low conflict scores. This visually confirms our core hypothesis that safety-utility conflicts are not global, but are localized to a small subset of specific heads.

    \item \textbf{Concentration in Middle-to-Deep Layers:} The ``Risky Zones'' tend to cluster in the middle and middle-deep layers (approximately layers 10 to 25). In contrast, the very shallow layers (input processing) and the final layers (output formatting) generally exhibit lower conflict scores. This suggests that the conflict between safety and general utility is most intense in the layers responsible for complex reasoning and semantic processing.

    \item \textbf{Intra-Layer Selectivity:} Even within the high-conflict layers, not all heads are risky. We typically observe only a few specific heads per layer showing high conflict intensity, while their neighboring heads remain safe. This further justifies the need for head-level diagnosis rather than layer-level pruning.
\end{itemize}

\paragraph{Observations and Analysis.}
Based on the visualization, we draw the following conclusions regarding the internal mechanics of alignment:

\begin{itemize}[leftmargin=*,itemsep=2pt, topsep=0pt, parsep=0pt]
    \item \textbf{Sparsity and Locality:} The conflict map is remarkably sparse. Only a small fraction of heads exhibit high conflict scores (the ``Risky Zone''), while the vast majority of heads remain in the ``Safe Zone.'' This explains why our sparse tuning strategy (updating only 25\% of heads) can be effective: the problematic parameters are few and localized.
    
    \item \textbf{Layer-Wise Distribution:} We observe that high-conflict heads often cluster in the \textbf{middle-to-late layers} of the Transformer. This aligns with mechanistic interpretability findings suggest that these layers are responsible for integrating semantic information and executing complex reasoning or refusal decisions.
    
    \item \textbf{Model-Specific Patterns:} While the general phenomenon of locality holds across Llama, Qwen, and Mistral, the specific ``hotspots'' differ. This underscores the necessity of \textbf{model-specific diagnosis} (as performed by CAST) rather than applying heuristic layer-range selection strategies.
\end{itemize}

\begin{table*}[h!]
\centering
\scriptsize
\setlength{\tabcolsep}{2.5pt}
\resizebox{\textwidth}{!}{
\begin{tabular}{c|l|c|c|cccc|ccc|cc|cc}
\toprule
\multirow{2}{*}{\textbf{Model}} & \multirow{2}{*}{\textbf{Method}} & \multirow{2}{*}{\textbf{Bucket}} & \textbf{Avg} & \multicolumn{4}{c|}{\textbf{General Capabilities}} & \multicolumn{3}{c|}{\textbf{Safety Performance}} & \multicolumn{2}{c|}{\textbf{Averages}} & \multicolumn{2}{c}{\textbf{Cost Ratios}} \\
& & & \textbf{Conf.} & MMLU & CSQA & GSM8K & MATH & AdvHarm & Wild & DAN & \textbf{Gen} & \textbf{Safe} & \textbf{UCR} & \textbf{MMLU-CR} \\ \midrule
\multirow{12}{*}{\rotatebox[origin=c]{90}{\textbf{Llama}}} 
 & \multirow{4}{*}{\shortstack{Unified\\Conflict}} 
 & $\mathcal{B}_1$ & 1.27 & 48.52 & 63.83 & 68.80 & 42.93 & 95.22 & 96.64 & 83.51 & 56.02 & 91.79 & 0.41 & 0.44 \\
 & & $\mathcal{B}_2$ & 0.88 & 52.40 & 66.28 & 69.33 & 41.07 & 95.92 & 97.44 & 80.52 & 57.27 & 91.29 & 0.37 & 0.29 \\
 & & $\mathcal{B}_3$ & 0.67 & 52.52 & 66.31 & 74.33 & 41.67 & 96.20 & 97.75 & 90.42 & 58.71 & 94.79 & 0.27 & 0.25 \\
 & & $\mathcal{B}_4$ & 0.47 & 55.73 & 69.42 & 77.20 & 43.00 & 94.85 & 96.69 & 86.33 & 61.34 & 92.62 & 0.19 & 0.14 \\ \cmidrule{2-15}
 & \multirow{4}{*}{\shortstack{Gradient\\Score}} 
 & $\mathcal{B}_1$ & 0.63 & 53.62 & 64.95 & 70.47 & 40.33 & 96.22 & 97.61 & 85.67 & 57.34 & 93.17 & 0.34 & 0.22 \\
 & & $\mathcal{B}_2$ & 0.52 & 55.49 & 68.53 & 70.13 & 43.47 & 96.52 & 97.39 & 85.27 & 59.40 & 93.06 & 0.26 & 0.15 \\
 & & $\mathcal{B}_3$ & 0.45 & 55.07 & 68.68 & 70.13 & 42.73 & 96.27 & 97.04 & 87.24 & 59.15 & 93.51 & 0.26 & 0.16 \\
 & & $\mathcal{B}_4$ & 0.33 & 53.09 & 65.74 & 79.67 & 43.40 & 96.25 & 97.84 & 89.42 & 60.48 & 94.50 & 0.21 & 0.23 \\ \cmidrule{2-15}
 & \multirow{4}{*}{\shortstack{Sensitivity\\Score}} 
 & $\mathcal{B}_1$ & 2.38 & 49.51 & 62.95 & 73.67 & 43.40 & 94.87 & 97.00 & 85.31 & 57.38 & 92.39 & 0.35 & 0.39 \\
 & & $\mathcal{B}_2$ & 1.86 & 51.61 & 66.01 & 77.07 & 41.00 & 95.85 & 97.13 & 85.31 & 58.92 & 92.76 & 0.28 & 0.30 \\
 & & $\mathcal{B}_3$ & 1.45 & 52.65 & 63.61 & 79.73 & 42.07 & 95.95 & 97.26 & 90.96 & 59.51 & 94.72 & 0.24 & 0.24 \\
 & & $\mathcal{B}_4$ & 1.13 & 56.93 & 69.61 & 71.27 & 42.80 & 96.00 & 97.35 & 88.85 & 60.15 & 94.07 & 0.22 & 0.09 \\ \midrule
\multirow{12}{*}{\rotatebox[origin=c]{90}{\textbf{Qwen}}} 
 & \multirow{4}{*}{\shortstack{Unified\\Conflict}} 
 & $\mathcal{B}_1$ & 1.22 & 65.68 & 77.23 & 36.20 & 40.20 & 71.85 & 84.75 & 75.37 & 54.83 & 77.32 & 1.10 & 0.15 \\
 & & $\mathcal{B}_2$ & 0.91 & 67.14 & 80.26 & 30.00 & 34.40 & 70.58 & 84.35 & 74.61 & 52.95 & 76.52 & 1.25 & 0.08 \\
 & & $\mathcal{B}_3$ & 0.71 & 67.51 & 80.84 & 29.87 & 37.40 & 74.52 & 86.52 & 79.15 & 53.90 & 80.06 & 1.01 & 0.05 \\
 & & $\mathcal{B}_4$ & 0.53 & 68.28 & 81.08 & 83.80 & 60.00 & 80.15 & 87.40 & 76.94 & 73.29 & 81.50 & 0.15 & 0.01 \\ \cmidrule{2-15}
 & \multirow{4}{*}{\shortstack{Gradient\\Score}} 
 & $\mathcal{B}_1$ & 0.57 & 66.78 & 79.88 & 29.53 & 35.13 & 69.08 & 84.88 & 78.96 & 52.83 & 77.64 & 1.19 & 0.09 \\
 & & $\mathcal{B}_2$ & 0.51 & 67.27 & 80.02 & 34.53 & 37.27 & 77.12 & 87.40 & 77.58 & 54.77 & 80.70 & 0.95 & 0.06 \\
 & & $\mathcal{B}_3$ & 0.48 & 66.82 & 81.00 & 28.53 & 34.40 & 65.05 & 81.30 & 70.63 & 52.69 & 72.33 & 1.61 & 0.12 \\
 & & $\mathcal{B}_4$ & 0.41 & 68.03 & 80.45 & 83.67 & 63.20 & 66.85 & 83.07 & 79.12 & 73.84 & 76.35 & 0.16 & 0.03 \\ \cmidrule{2-15}
 & \multirow{4}{*}{\shortstack{Sensitivity\\Score}} 
 & $\mathcal{B}_1$ & 2.38 & 66.17 & 79.55 & 28.87 & 34.60 & 77.05 & 85.10 & 74.64 & 52.30 & 78.93 & 1.14 & 0.11 \\
 & & $\mathcal{B}_2$ & 1.86 & 67.46 & 80.56 & 23.07 & 30.13 & 71.68 & 85.06 & 73.67 & 50.31 & 76.80 & 1.37 & 0.06 \\
 & & $\mathcal{B}_3$ & 1.45 & 67.07 & 79.50 & 55.47 & 52.73 & 71.67 & 87.67 & 78.80 & 63.69 & 79.38 & 0.60 & 0.07 \\
 & & $\mathcal{B}_4$ & 1.13 & 65.71 & 80.18 & 29.60 & 39.00 & 78.38 & 86.25 & 75.66 & 53.62 & 80.10 & 1.02 & 0.13 \\ \midrule
\multirow{12}{*}{\rotatebox[origin=c]{90}{\textbf{Mistral}}} 
 & \multirow{4}{*}{\shortstack{Unified\\Conflict}} 
 & $\mathcal{B}_1$ & 1.21 & 27.78 & 25.72 & 19.00 & 4.60 & 60.35 & 81.56 & 68.75 & 19.27 & 70.22 & 0.81 & 0.95 \\
 & & $\mathcal{B}_2$ & 0.89 & 41.81 & 45.21 & 18.20 & 5.80 & 66.25 & 78.79 & 68.33 & 27.76 & 71.12 & 0.37 & 0.25 \\
 & & $\mathcal{B}_3$ & 0.68 & 45.17 & 54.71 & 19.60 & 5.40 & 53.25 & 74.40 & 60.36 & 31.22 & 62.67 & 0.34 & 0.14 \\
 & & $\mathcal{B}_4$ & 0.46 & 47.62 & 56.35 & 18.80 & 7.40 & 54.35 & 77.59 & 63.49 & 32.54 & 65.14 & 0.19 & 0.00 \\ \cmidrule{2-15}
 & \multirow{4}{*}{\shortstack{Gradient\\Score}} 
 & $\mathcal{B}_1$ & 0.61 & 37.19 & 36.28 & 22.20 & 6.60 & 61.14 & 78.24 & 65.28 & 25.57 & 68.22 & 0.55 & 0.54 \\
 & & $\mathcal{B}_2$ & 0.51 & 30.69 & 32.92 & 24.40 & 6.19 & 52.81 & 77.71 & 65.48 & 23.55 & 65.33 & 0.78 & 1.07 \\
 & & $\mathcal{B}_3$ & 0.45 & 43.91 & 39.65 & 25.60 & 9.60 & 50.60 & 74.39 & 61.50 & 29.69 & 62.16 & 0.48 & 0.25 \\
 & & $\mathcal{B}_4$ & 0.34 & 46.86 & 54.39 & 24.00 & 7.00 & 61.30 & 77.59 & 65.20 & 33.06 & 68.03 & 0.13 & 0.01 \\ \cmidrule{2-15}
 & \multirow{4}{*}{\shortstack{Sensitivity\\Score}} 
 & $\mathcal{B}_1$ & 2.35 & 42.26 & 48.48 & 21.40 & 7.00 & 52.69 & 78.65 & 61.64 & 29.78 & 64.33 & 0.40 & 0.33 \\
 & & $\mathcal{B}_2$ & 1.85 & 46.26 & 53.16 & 18.80 & 8.81 & 55.71 & 76.91 & 64.90 & 31.76 & 65.84 & 0.24 & 0.05 \\
 & & $\mathcal{B}_3$ & 1.44 & 49.87 & 62.58 & 22.00 & 7.20 & 55.30 & 77.72 & 62.99 & 35.41 & 65.34 & 0.00 & 0.00 \\
 & & $\mathcal{B}_4$ & 1.12 & 47.63 & 56.35 & 18.80 & 7.40 & 54.35 & 77.59 & 63.49 & 32.55 & 65.14 & 0.19 & 0.00 \\ \bottomrule
\end{tabular}
}
\caption{\textbf{Source Data for Predictive Correlation Analysis.} Detailed breakdown of the Average Conflict Score versus realized utility and safety outcomes across all buckets ($\mathcal{B}_1$-$\mathcal{B}_4$) and models. This data underpins the UCR and MMLU-CR correlations reported in Section~\ref{sec:rq3}.}
\label{tab:full_correlation_data}
\end{table*}
\section{Detailed Experimental Results for Correlation Analysis}
\label{app:correlation_results}

In Section~\ref{sec:rq3}, we visualized the correlation between the Conflict Score and the realized Utility Cost Ratio (UCR). Table~\ref{tab:full_correlation_data} provides the raw numerical data supporting these visualizations. 

We report the exact Avg Conflict Score alongside fine-grained performance metrics (General Utility and Safety) for every experimental bucket across three architectures: Llama, Qwen, and Mistral. These values are used to calculate the UCR and MMLU-CR, verifying that higher conflict scores consistently correspond to steeper utility penalties (higher UCR) and lower general capabilities.

\section{Additional Ablation Studies}
\label{app:ablation}
In this section, we provide the comprehensive experimental results supporting the robustness analysis summarized in the main text. We rigorously validate the core design principles of CAST across four critical dimensions: the necessity of the unified metric, sensitivity to sparsity ratios, robustness to data properties, and computational complexity.

\begin{table}[t]
\centering
\resizebox{1.0\columnwidth}{!}{
\begin{tabular}{cc|cc|cc}
\toprule
\multirow{2}{*}{\textbf{Model}} & \multirow{2}{*}{\textbf{Metric}} & \multicolumn{2}{c|}{\textbf{U-CR}} & \multicolumn{2}{c}{\textbf{MMLU-CR}} \\
\cmidrule(lr){3-4} \cmidrule(lr){5-6}
 & & \textbf{Pearson} & \textbf{Spearman} & \textbf{Pearson} & \textbf{Spearman} \\
\midrule

\multirow{3}{*}{Llama} 
 & $\mathcal{O}$ & 0.96 & 0.80 & -0.13 & -0.40 \\
 & $\mathcal{S}$ & \textbf{0.99} & \textbf{1.00} & 0.96 & \textbf{1.00} \\
 & \boldmath$\mathcal{C}$ & 0.95 & \textbf{1.00} & \textbf{0.99} & \textbf{1.00} \\
\midrule

\multirow{3}{*}{Qwen} 
 & $\mathcal{O}$ & 0.62 & 0.40 & 0.52 & 0.40 \\
 & $\mathcal{S}$ & 0.46 & 0.60 & -0.10 & -0.40 \\
 & \boldmath$\mathcal{C}$ & \textbf{0.73} & \textbf{0.80} & \textbf{1.00} & \textbf{1.00} \\
\midrule

\multirow{3}{*}{Mistral} 
 & $\mathcal{O}$ & 0.77 & 0.80 & 0.67 & 0.80 \\
 & $\mathcal{S}$ & 0.73 & 0.80 & 0.85 & 0.80 \\
 & \boldmath$\mathcal{C}$ & \textbf{0.95} & \textbf{1.00} & \textbf{0.94} & \textbf{1.00} \\

\bottomrule
\end{tabular}
}
\caption{\small \textbf{Predictive Validity of Diagnostic Components.} Evaluation of Gradient-only ($\mathcal{O}$), Sensitivity-only ($\mathcal{S}$), and our Unified metric ($\mathcal{C}$). The unified metric consistently correlates with realized utility costs across all models.}
\label{tab:ablation_correlation_full}
\end{table}
\begin{table*}[t]
\centering
\small
\setlength{\tabcolsep}{4pt}
\begin{tabular}{l|cccc|ccc|cc}
\toprule
\multirow{2}{*}{\textbf{Sparsity Ratio}} & \multicolumn{4}{c|}{\textbf{General Utility}} & \multicolumn{3}{c|}{\textbf{Safety Performance}} & \multicolumn{2}{c}{\textbf{Average}} \\
\cmidrule(lr){2-5} \cmidrule(lr){6-8} \cmidrule(lr){9-10}
& MMLU & CSQA & GSM8K & MATH & AdvHarm & WildGuard & DAN & \textbf{Gen Avg} & \textbf{Safe Avg} \\
\midrule
Full SFT (100\%) & 46.28 & 60.84 & 20.80 & 19.20 & 93.26 & 94.29 & 84.27 & 36.78 & 90.61 \\
\midrule
\multicolumn{10}{c}{\textit{Risky Zone Trajectory (Prioritizing High-Conflict Heads)}} \\
\midrule
Top-25\% & 48.52 & 63.83 & 68.80 & 42.93 & 95.22 & 96.64 & 83.51 & 56.02 & 91.79 \\
Top-50\% & 47.79 & 58.72 & 70.60 & 42.00 & 96.30 & 97.61 & 89.54 & 54.78 & 94.48 \\
Top-75\% & 50.98 & 67.16 & 73.20 & 40.00 & 96.75 & 97.35 & 90.11 & 57.83 & 94.73 \\
\midrule
\multicolumn{10}{c}{\textit{Safe Zone Trajectory (CAST, Prioritizing Low-Conflict Heads)}} \\
\midrule
\textbf{Bottom-25\% (Ours)} & \textbf{55.73} & \textbf{69.42} & \textbf{77.20} & \textbf{43.00} & 94.85 & 96.69 & 86.33 & \textbf{61.34} & 92.62 \\
Bottom-50\% & 53.67 & 67.65 & 68.00 & 42.40 & 94.90 & 95.23 & 86.33 & 57.93 & 92.15 \\
Bottom-75\% & 53.24 & 65.85 & 61.80 & 43.00 & \textbf{96.70} & \textbf{97.61} & \textbf{90.89} & 55.97 & \textbf{95.07} \\
\bottomrule
\end{tabular}
\caption{\textbf{Sensitivity Analysis to Sparsity Ratio on Llama.} We compare the performance trajectory when prioritizing High-Conflict heads (Top-k\%) vs. Low-Conflict heads (Bottom-k\%). The Bottom-25\% setting achieves the optimal trade-off, maximizing general utility while maintaining robust safety.}
\label{tab:sparsity_ablation}
\end{table*}
\begin{table*}[t]
\centering
\small
\resizebox{\textwidth}{!}{
\begin{tabular}{c|l|cccc|c|ccc|c}
\toprule
\multirow{2}{*}{\textbf{Model}} & \multirow{2}{*}{\textbf{Bucket}} & \multicolumn{4}{c|}{\textbf{General Capabilities}} & \textbf{Avg.} & \multicolumn{3}{c|}{\textbf{Safety Performance}} & \textbf{Avg.} \\
\cmidrule(lr){3-6} \cmidrule(lr){8-10}
 & & MMLU & CSQA & GSM8K & MATH & \textbf{General} & AdvHarm & WildGuard & DAN & \textbf{Safety} \\ \midrule
\multirow{4}{*}{\shortstack{\textbf{Llama}\\(MMLU-100)}} 
 & $\mathcal{B}_1$ (High C) & 50.61 & 64.62 & 66.00 & 41.20 & 55.61 & 96.45 & 97.88 & 88.90 & 94.41 \\
 & $\mathcal{B}_2$ & 50.63 & 67.32 & 71.60 & 42.20 & 57.94 & 97.15 & 98.28 & 92.88 & \textbf{96.10} \\
 & $\mathcal{B}_3$ & 54.27 & 64.95 & 73.80 & 43.20 & 59.05 & 95.15 & 96.42 & 80.07 & 90.55 \\
 & \cellcolor{mygreen}$\mathcal{B}_4$ (Low C) & \cellcolor{mygreen}\textbf{56.89} & \cellcolor{mygreen}\textbf{70.43} & \cellcolor{mygreen}\textbf{81.00} & \cellcolor{mygreen}\textbf{44.40} & \cellcolor{mygreen}\textbf{63.18} & \cellcolor{mygreen}94.10 & \cellcolor{mygreen}97.35 & \cellcolor{mygreen}85.20 & \cellcolor{mygreen}92.21 \\ \midrule
\multirow{4}{*}{\shortstack{\textbf{Qwen}\\(MMLU-100)}} 
 & $\mathcal{B}_1$ (High C) & 68.00 & 81.16 & \textbf{86.80} & 59.60 & 73.89 & 83.80 & 89.12 & 81.00 & 84.64 \\
 & $\mathcal{B}_2$ & 68.09 & 81.08 & 58.80 & 51.80 & 64.94 & \textbf{88.60} & \textbf{89.79} & \textbf{81.07} & \textbf{86.49} \\
 & $\mathcal{B}_3$ & 68.37 & 81.16 & 80.80 & 60.60 & 72.73 & 69.30 & 82.89 & 78.58 & 76.92 \\
 & \cellcolor{mygreen}$\mathcal{B}_4$ (Low C) & \cellcolor{mygreen}\textbf{68.47} & \cellcolor{mygreen}\textbf{81.57} & \cellcolor{mygreen}80.40 & \cellcolor{mygreen}\textbf{68.40} & \cellcolor{mygreen}\textbf{74.71} & \cellcolor{mygreen}67.70 & \cellcolor{mygreen}82.76 & \cellcolor{mygreen}76.94 & \cellcolor{mygreen}75.80 \\ \bottomrule
\end{tabular}
}
\caption{\small
\textbf{Data Efficiency Analysis (MMLU-100 Calibration).} We report full metrics for conflict diagnosis using only 100 MMLU samples. Across both Llama and Qwen, the Low-Conflict bucket ($\mathcal{B}_4$) consistently identifies the optimal region for General Utility, validating the robustness of the metric under sparse data conditions.}
\label{tab:data_efficiency}
\end{table*}
\begin{table*}[!t]
\centering
\small
\resizebox{\textwidth}{!}{
\begin{tabular}{c|l|cccc|c|ccc|c}
\toprule
\multirow{2}{*}{\textbf{Model}} & \multirow{2}{*}{\textbf{Bucket}} & \multicolumn{4}{c|}{\textbf{General Capabilities}} & \textbf{Avg.} & \multicolumn{3}{c|}{\textbf{Safety Performance}} & \textbf{Avg.} \\
\cmidrule(lr){3-6} \cmidrule(lr){8-10}
 & & MMLU & CSQA & GSM8K & MATH & \textbf{General} & AdvHarm & WildGuard & DAN & \textbf{Safety} \\ \midrule
\multirow{4}{*}{\shortstack{\textbf{Llama}\\(GSM8K-100)}} 
 & $\mathcal{B}_1$ (High C) & 56.52 & 69.37 & 70.20 & 41.60 & 59.42 & 96.40 & 97.08 & 86.62 & 93.37 \\
 & $\mathcal{B}_2$ & 53.64 & 65.03 & 72.60 & 42.40 & 58.42 & 96.25 & 96.68 & \textbf{93.74} & \textbf{95.56} \\
 & \cellcolor{mygreen}$\mathcal{B}_3$ (Optimal) & \cellcolor{mygreen}54.66 & \cellcolor{mygreen}\textbf{68.55} & \cellcolor{mygreen}\textbf{83.40} & \cellcolor{mygreen}41.60 & \cellcolor{mygreen}\textbf{62.05} & \cellcolor{mygreen}\textbf{96.55} & \cellcolor{mygreen}\textbf{98.54} & \cellcolor{mygreen}89.75 & \cellcolor{mygreen}94.95 \\
 & $\mathcal{B}_4$ (Low C) & 51.72 & 67.32 & 63.40 & 36.40 & 54.71 & 95.85 & 96.29 & 87.83 & 93.32 \\ \midrule
\multirow{4}{*}{\shortstack{\textbf{Qwen}\\(GSM8K-100)}} 
 & $\mathcal{B}_1$ (High C) & 68.26 & 81.74 & 75.40 & 66.80 & 73.05 & 77.40 & 88.20 & 74.16 & 79.92 \\
 & \cellcolor{mygreen}$\mathcal{B}_2$ (Optimal) & \cellcolor{mygreen}\textbf{68.94} & \cellcolor{mygreen}81.08 & \cellcolor{mygreen}\textbf{86.60} & \cellcolor{mygreen}\textbf{67.40} & \cellcolor{mygreen}\textbf{76.00} & \cellcolor{mygreen}82.15 & \cellcolor{mygreen}88.59 & \cellcolor{mygreen}\textbf{78.22} & \cellcolor{mygreen}82.99 \\
 & $\mathcal{B}_3$ & 67.36 & \textbf{81.82} & 44.60 & 51.00 & 61.19 & \textbf{86.30} & \textbf{89.79} & 78.15 & \textbf{84.75} \\
 & $\mathcal{B}_4$ (Low C) & 67.74 & 80.51 & 80.20 & 61.80 & 72.56 & 73.80 & 86.34 & 74.66 & 78.27 \\ \bottomrule
\end{tabular}
}
\caption{\small
\textbf{Domain Transfer Analysis (GSM8K-100 Calibration).} We switch the diagnostic source to 100 samples from GSM8K. Unlike the monotonic trend in MMLU calibration, both Llama and Qwen exhibit an ``Inverted U-shape'' or intermediate peak ($\mathcal{B}_3$ and $\mathcal{B}_2$, respectively), confirming that reasoning tasks require a specific trade-off zone distinct from the extreme tail.}
\label{tab:domain_transfer}
\end{table*}

\subsection{Necessity of Coupled Diagnosis}
To validate the necessity of coupling \textit{Optimization Conflict} ($\mathcal{O}$) and \textit{Functional Sensitivity} ($\mathcal{S}$), we compare our unified metric $C(h)$ against two decoupled variants: \textbf{Gradient-Only} ($\mathcal{O}$) and \textbf{Sensitivity-Only} ($\mathcal{S}$). We quantify predictive power via the Pearson correlation ($r$) with the post-alignment Utility Cost Ratio (UCR).

As presented in Table~\ref{tab:ablation_correlation_full}, decoupled metrics exhibit significant instability. While single components may capture risk in specific contexts (e.g., $\mathcal{S}$ on Llama), they suffer from critical blind spots. For instance, $\mathcal{S}$ fails to predict MMLU degradation in Qwen ($r = -0.10$), and $\mathcal{O}$ negatively correlates with MMLU-CR in Llama ($r = -0.13$).
By contrast, our unified score $C(h)$ demonstrates universal robustness, maintaining strong positive correlations ($r \in [0.73, 1.00]$) across all models. This confirms that accurate risk diagnosis requires the intersection of both geometric adversariality and functional sensitivity.

\subsection{Sensitivity to Sparsity Ratio}
\label{sec:sparsity_sensitivity}

In the main paper, we adopted a fixed sparsity ratio of 25\% for our primary experiments. To justify this choice and verify the robustness of our \textbf{Conflict Score} ranking across different budget constraints, we conduct a detailed sensitivity analysis \textbf{on the representative Llama model} by varying the percentage of trainable heads from 25\% to 100\%.

\paragraph{Setup.} We define two contrasting selection trajectories based on the conflict ranking $C(h)$:
\begin{enumerate}
    \item \textbf{CAST (Safe Zone) Trajectory:} We progressively include heads starting from the \textit{lowest} conflict scores (Bottom-25\% $\rightarrow$ 100\%). This tests the efficacy of prioritizing safety-agnostic parameters.
    \item \textbf{Risky Zone Trajectory:} We progressively include heads starting from the \textit{highest} conflict scores (Top-25\% $\rightarrow$ 100\%), representing the worst-case scenario.
\end{enumerate}

\paragraph{Results and Analysis.}
Table~\ref{tab:sparsity_ablation} presents the detailed performance across all ratios. We observe three critical phenomena:

\begin{itemize}[leftmargin=*]
    \item \textbf{Consistent Pareto Dominance:} The CAST (Safe Zone) selection consistently outperforms the Risky Zone. At the 25\% budget, the Bottom-25\% selection achieves a General Average of \textbf{61.34}, significantly beating the Top-25\% (56.02). This gap is most visible in reasoning tasks like GSM8K, where the Safe Zone maintains performance at \textbf{77.20}, whereas the Risky Zone drops to 68.80.

    \item \textbf{Optimal Sweet Spot at 25\%:} Contrary to the intuition that ``more parameters equal better performance,'' we see that adding more heads to the Safe Zone actually hurts utility. Increasing the ratio from Bottom-25\% to Bottom-50\% drops the General Average from \textbf{61.34} to \textbf{57.93}, without improving safety. This suggests that the Top-25\% of Low-Conflict heads are sufficient for alignment, and adding more simply introduces noise.

    \item \textbf{The ``Full SFT'' Cliff:} Interestingly, within the Risky Zone, performance recovers slightly as we expand from Top-25\% to Top-75\% (MMLU rises to 50.98), likely because more low-conflict heads are included. However, this trend ends abruptly at 100\% (Full SFT), where the General Average crashes to \textbf{36.78}. This sharp drop confirms that blindly updating every parameter triggers the worst-case conflict, far exceeding the harm of updating just the risky subset.
\end{itemize}

\subsection{Sensitivity to Data Size: Data Efficiency}
\label{sec:data_efficiency}

To assess whether our conflict diagnosis requires large-scale data, we restrict the calibration set to only \textbf{100 samples} from MMLU (a $5\times$ reduction). We evaluate this setting on both Llama and Qwen to verify cross-model consistency.

\paragraph{Results.} As detailed in Table~\ref{tab:data_efficiency}, despite the minimal data, the conflict metric remains highly effective in identifying the optimal utility zone.
\begin{itemize}
    \item \textbf{Llama:} We observe a strictly monotonic trend. The General Average improves progressively from 55.61\% in the Risky Zone ($\mathcal{B}_1$) to \textbf{63.18\%} in the Safe Zone ($\mathcal{B}_4$). Notably, $\mathcal{B}_4$ achieves the best performance across all individual utility tasks (MMLU, CSQA, GSM8K, MATH).
    \item \textbf{Qwen:} Similarly, the Low-Conflict bucket ($\mathcal{B}_4$) yields the highest General Average of \textbf{74.71\%}, significantly outperforming the intermediate buckets ($\mathcal{B}_2$: 64.94\%).
\end{itemize}
This confirms that the "Conflict Landscape" for general knowledge is a salient structural property that can be accurately captured even with extremely sparse signals, ensuring the method's efficiency and reproducibility across different architectures.

\subsection{Sensitivity to Data Source: The ``Inverted U-Shape'' Phenomenon}
\label{sec:domain_transfer}

While MMLU calibration yields consistent trends favoring the "Low Conflict" tail, we further investigate whether these conflict maps transfer across domains. We replace the calibration source with \textbf{100 samples from GSM8K} to diagnose conflict specifically for reasoning tasks.

\paragraph{Results.} As presented in Table~\ref{tab:domain_transfer}, switching the data source reveals a distinct \textbf{``Inverted U-shape''} pattern, where the optimal trade-off shifts from the tail ($\mathcal{B}_4$) to intermediate buckets.
\begin{itemize}
    \item \textbf{Llama:} The utility peak appears at $\mathcal{B}_3$ (\textbf{62.05\%} Gen Avg), with GSM8K accuracy reaching \textbf{83.40\%}. This significantly outperforms the Low-Conflict tail $\mathcal{B}_4$ (GSM8K: 63.40\%, Gen Avg: 54.71\%).
    \item \textbf{Qwen:} A similar structural "Sweet Spot" is observed at $\mathcal{B}_2$. It achieves the highest General Average (\textbf{76.00\%}) and GSM8K score (\textbf{86.60\%}), surpassing the $\mathcal{B}_4$ bucket (72.56\% and 80.20\% respectively).
\end{itemize}

\paragraph{Analysis: The Limitation of Single-Head Ablation.}
We attribute this ``Inverted U-shape'' to the limitations of using single-head ablation to measure sensitivity in reasoning tasks. Unlike factual queries, reasoning relies on complex, multi-step logical chains. Masking a single head often fails to break this chain due to model redundancy, leading our metric $\mathcal{S}(h)$ to underestimate the head's importance. Consequently, many critical reasoning heads are incorrectly assigned low scores and grouped into the ``Low Conflict'' bucket ($\mathcal{B}_4$). However, when these heads are modified during fine-tuning, the subtle changes disrupt the precise logic flow. As a result, the true ``safe zone'' shifts to intermediate buckets, where the balance between safety and valid sensitivity estimation is better maintained.

\subsection{Computational Complexity}
Benchmarked on Llama (NVIDIA H800), analyzing a single head (100 samples) takes approximately 7 seconds for utility (next-token prediction) and 1 minute for safety (generation). For a model with $N$ heads, the total time scales linearly to $N \times \text{Cost}_{\text{head}}$. While this represents a fixed upfront cost, it is a strict \textbf{one-off pre-computation}. Unlike random search strategies that require expensive, repeated training runs to stumble upon an effective sparsity pattern, our approach generates a precise Conflict Map in a single pass. This makes CAST not only more interpretable but also significantly more cost-effective than blind trial-and-error.

\end{document}